\documentclass[runningheads]{llncs}
\usepackage{graphicx}
\usepackage{tikz}
\usepackage{comment}
\usepackage{amsmath,amssymb} % define this before the line numbering.
\usepackage{color}
\usepackage[flushleft]{threeparttable}

\usepackage{xcolor}
\usepackage{color, colortbl}
\usepackage{booktabs}
\usepackage{subfigure}
\usepackage{float}

\definecolor{Gray}{gray}{0.9}
\definecolor{Red}{RGB}{230, 57, 70}
\definecolor{Blue}{RGB}{0, 100, 148}

\newcommand{\etal}{\textit{et al}.\@ }
\newcommand{\name}{\emph{DeciWatch}\xspace}

\usepackage{mmstyle}
\usepackage{wrapfig}

\begin{document}
% \linenumbers
\pagestyle{headings}
\mainmatter
\def\ECCVSubNumber{100}  % Insert your submission number here

\title{DeciWatch: A Simple Baseline for $10\times$ Efficient 2D and 3D Pose Estimation} % Replace with your title

\begin{comment}
\titlerunning{ECCV-22 submission ID \ECCVSubNumber} 
\authorrunning{ECCV-22 submission ID \ECCVSubNumber} 
\author{Anonymous ECCV submission}
\institute{Paper ID \ECCVSubNumber}
\end{comment}
%******************

%\begin{comment}
\titlerunning{DeciWatch}
\author{Ailing Zeng$^{1}$  \and
Xuan Ju$^{1}$  \and
Lei Yang$^{2}$  \and
Ruiyuan Gao$^{1}$  \and
Xizhou Zhu$^{2}$  \and
Bo Dai$^{3}$  \and
Qiang Xu$^{1}$ 
}
\authorrunning{A. Zeng et al.}
\institute{$^{1}$The Chinese University of Hong Kong, $^{2}$Sensetime Group Limited, \\$^{3}$Shanghai AI Laboratory\\
\email{\{alzeng, qxu\}@cse.cuhk.edu.hk}
}
%\end{comment}
%******************

\maketitle

\begin{abstract}

This paper proposes a simple baseline framework for video-based 2D/3D human pose estimation that can achieve $10\times$ efficiency improvement over existing works without any performance degradation, named \name. Unlike current solutions that estimate each frame in a video, \name introduces a simple yet effective \emph{sample-denoise-recover} framework that only watches sparsely sampled frames, taking advantage of the continuity of human motions and the lightweight pose representation. Specifically, \name uniformly samples less than $10\%$ video frames for detailed estimation, denoises the estimated 2D/3D poses with an efficient Transformer architecture, and then accurately recovers the rest of the frames using another Transformer-based network. Comprehensive experimental results on three video-based human pose estimation, body mesh recovery tasks and efficient labeling in videos with four datasets validate the efficiency and effectiveness of \name. Code is available at \url{https://github.com/cure-lab/DeciWatch}.

%\vspace{-5pt}
%\vspace{-5pt}
\end{abstract}
\section{Introduction}\label{introduction}
% \vspace{-0.2cm}
2D/3D human pose estimation~\cite{liu2021recent,desmarais2021review,zheng2020deep} has numerous applications, such as surveillance, virtual reality, and autonomous driving. Various high-performance image-based pose estimators~\cite{newell2016stacked,xiao2018simple,sun2019deep,joo2020eft,kolotouros2019spin,kocabas2021pare} are proposed in the literature, but they are associated with substantial computational costs. 

\begin{figure}[t!]	
\centering
 	\subfigure[Compact Network Design for Pose Estimation] 
 	{
 		\begin{minipage}[t]{0.98\linewidth}
 			\centering         
 			\includegraphics[width=4.5in]{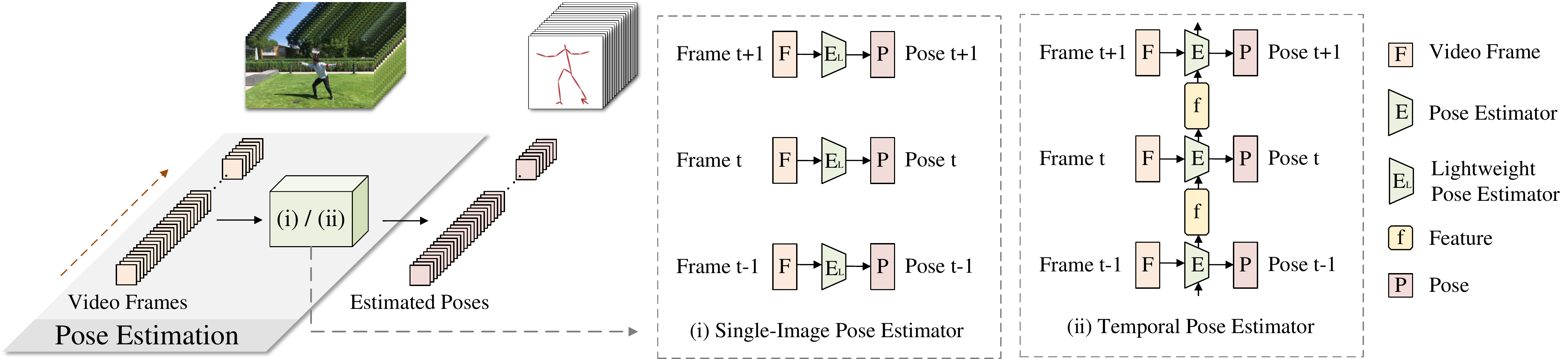}   
 		\end{minipage}
 		\label{fig:framework-a} 
 	} 
     	
 	\subfigure[Keyframe-Based Efficient Pose Estimation] 
 	{
		\begin{minipage}[t]{0.98\linewidth}
			\centering      
			\includegraphics[width=4.5in]{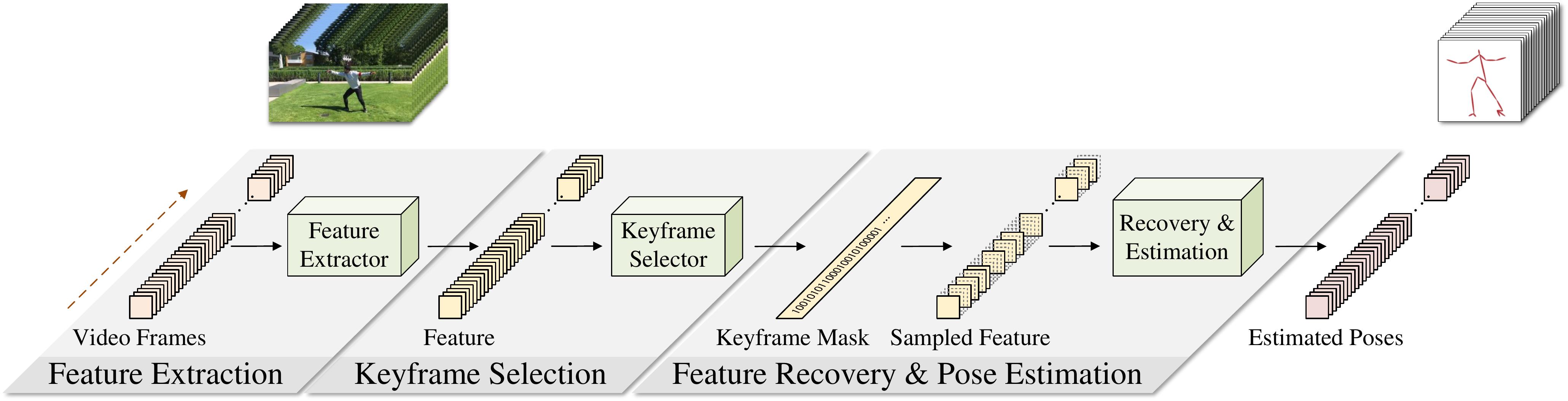}
		\end{minipage}
		\label{fig:framework-b} 
	}
	
	\subfigure[Our Sample-Denoise-Recover Framework (\name)] 
 	{
		\begin{minipage}[t]{0.98\linewidth}
			\centering      
			\includegraphics[width=4.5in]{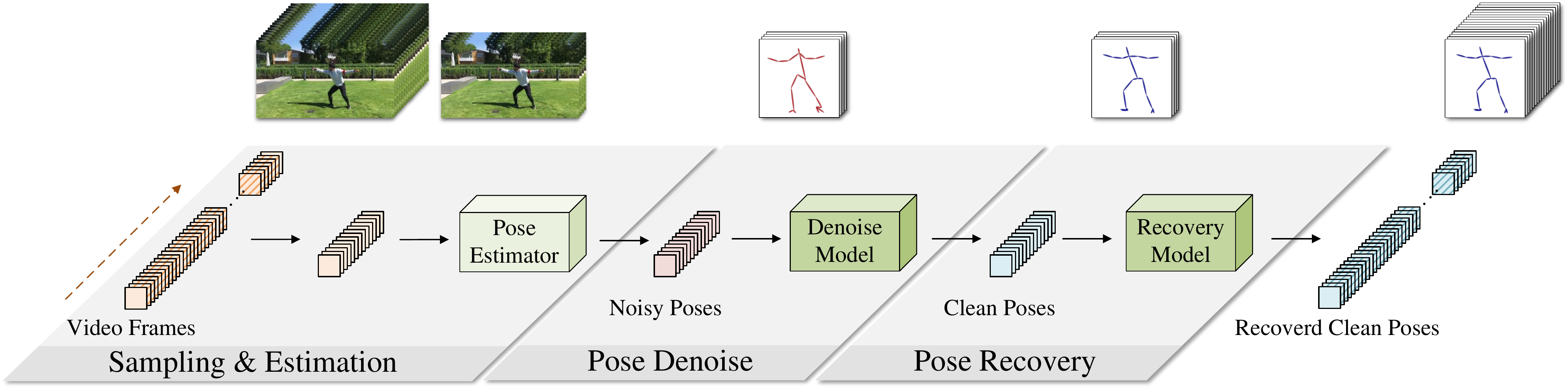}
		\end{minipage}
		\label{fig:framework-c} 
	}
    	
% \vspace{-0.1cm}
\caption{The workflows of three types of efficient pose estimation frameworks. (a) is compact model designs. The (green) pose estimation module has two design strategies: (\romannumeral1) shows single-frame efficient methods~\cite{cao2019openpose,li2021online,hwang2020lightweight,zhao2021estimating,dai2021fasterpose,zhang2019simple,osokin2018real,zheng2021lightweight,choi2021mobilehumanpose,hinton2015distilling} that use lightweight models to reduce the costs of each frame; (\romannumeral2) presents some temporal efficient strategies~\cite{fan2020adaptive,nie2019dynamic} that utilize feature similarities among consecutive frames via RNNs to decrease feature extraction cost. (b) is the keyframe-based efficient framework\cite{zhang2020key,fan2021motion}. They first select about 30\%$\sim$40\% keyframes in a video by watching all frames, then recover the whole sequence based on features of selected keyframes. (c) is the proposed efficient \emph{sample-denoise-recover} framework \name with 5\%$\sim$10\% frames watched.}
\label{fig:three_framewors} 
%\vspace{-0.4cm}
\end{figure}

There are two main approaches to improving the efficiency of human pose estimators so that they can be deployed on resource-scarce edge devices (e.g., smart cameras). A straightforward way to improve the efficiency is designing more compact models, such as numerous light-weighted image-level pose estimators \cite{cao2019openpose,li2021online,hwang2020lightweight,zhao2021estimating,dai2021fasterpose,zhang2019simple,osokin2018real,zheng2021lightweight,choi2021mobilehumanpose,hinton2015distilling,yu2021lite} (see Fig.~\ref{fig:framework-a}(\romannumeral1)) and video-level pose estimators \cite{nie2019dynamic,fan2020adaptive} (see Fig.~\ref{fig:framework-a}(\romannumeral2)) introduced in previous literature. 
However, when estimating on a video, such approaches inevitably lead to a sub-optimal solution for efficiency improvement due to the frame-by-frame estimation scheme.
In contrast, a promising but rarely explored direction is leveraging the semantic redundancy among frames of videos,
where we can feed only keyframes to heavy and high-performance modules and recover or estimate the rest of the frames with light-weighted modules \cite{zhang2020key,fan2021motion} (see Fig.~\ref{fig:framework-b}).
While the computational efficiency of these works is improved due to the use of keyframes, they still need to conduct costly feature extraction on each frame for keyframe selection, making it hard to further reduce their computational complexity.

To achieve highly efficient 2D/3D pose estimation \emph{without the need of watching every frame in a video},
we propose a novel framework based on the continuity of human motions, which conducts pose estimation only on sparsely sampled video frames. 
Since these detected poses ineluctably contain various noises, they will affect the effectiveness of the recovery. 
Subsequently, poses of those sampled frames should be denoised before recovered, where we formulate the three-step \emph{sample-denoise-recover} framework.
By doing so, the problem in the \emph{recover} stage is similar to the long-standing motion completion task in the computer graphics literature~\cite{kucherenko2018neural,ji2020missing,skurowski2021gap,kaufmann2020convolutional,cai2021unified,duan2021single}. However, there are two main differences: (\romannumeral1). our objective is to achieve highly efficient pose estimation, and hence we could only afford lightweight models for pose recovery on frames that are not processed by pose estimators; (\romannumeral2). most existing motion completion works assume ground-truth poses on the given keyframes. In contrast, the visible frames in our task could have untrustworthy poses with challenging occlusion or rarely seen actions. 
%The validity of the motion completion solutions on our benchmarks is unexplored.}

This work proposes a simple yet effective baseline framework (see Fig.~\ref{fig:framework-c}) that watches sparsely sampled frames for highly efficient 2D and 3D video-based human pose estimation. We empirically show that we could maintain and even improve the pose estimation accuracy, with less than $10\%$ frames calculated with the costly pose estimator. We name the proposed framework \name, and the contributions of this work include:

% \vspace{-0.1cm}
\begin{itemize}

\item To the best of our knowledge, this is the first work that considers sparsely sampled frames in video-based pose estimation tasks. \name is compatible with any given single-frame pose estimator, achieving $10\times$ efficiency improvement without any performance degradation. Moreover, the pose sequence obtained by \name is much smoother than existing solutions as it naturally models the continuity of human motions. 
\item We propose a novel sample-denoise-recover pipeline in \name. Specifically, we uniformly sample less than $10\%$ of video frames for estimation, denoise the estimated 2D/3D poses with an efficient Transformer architecture named \emph{DenoiseNet}, and then accurately recover the poses for the rest of the frames using another Transformer network called \emph{RecoverNet}. Thanks to the lightweight pose representation, the two subnets in our design are much smaller than the costly pose estimator. 
\item We verify the efficiency and effectiveness of \name on three human pose estimation, body recovery tasks, and efficient labeling in videos with four widely-used datasets and five popular single-frame pose estimators as  backbones. We also conduct extensive ablation studies and point out future research directions to further enhance video-based tasks' efficiency. 
%We also present some interesting analyses to facilitate future research.
\end{itemize}
% \vspace{-0.1cm}

\section{Related Work}\label{related_work}
%\vspace{-0.2cm}
\subsection{Efficient Human Pose Estimation}
\label{sec:related_efficient}
%\vspace{-0.1cm}
% Human Pose Estimation is the task of estimating human joints or meshes from images. 
Efficient attempts at human pose estimation can be divided into image-based and video-based.
Image-based efficient pose estimators~\cite{cao2019openpose,li2021online,hwang2020lightweight,zhao2021estimating,dai2021fasterpose,zhang2019simple,osokin2018real,zheng2021lightweight,choi2021mobilehumanpose,yu2021lite} mainly focus on employing well-designed network structures~\cite{cao2019openpose,zhang2019simple,osokin2018real,zheng2021lightweight,choi2021mobilehumanpose,yu2021lite,zeng2020srnet}, knowledge distillation~\cite{li2021online,hwang2020lightweight,hinton2015distilling}, or low-resolution features~\cite{zhao2021estimating,dai2021fasterpose,li2021rle} to reduce model capacity and decrease spatial redundancies, where they may suffer from accuracy reduction, especially in the cases of complex and rare poses.
% These methods take advantage of spatial redundancies and therefore can improve efficiency while maintaining accuracy. 
Moreover, when dealing with videos, these methods reveal their limitations for having to estimate poses frame-by-frame. Their outputs also suffer from unavoidable jitters because they lack the capability of using temporal information.

To cope with video inputs, other attempts exploit temporal co-dependency among consecutive frames to decrease unnecessary calculations. However, only a few video-based efficient estimation methods~\cite{fan2020adaptive,nie2019dynamic,zhang2020key,fan2021motion} are proposed in the literature, and they mainly target on 2D pose estimation. In particular, 
%ACE~\cite{fan2020adaptive} extracts feature dynamically between a lightweight model and a heavy network via a Gaussian kernel based gate module for 3D hand pose estimation. 
DKD~\cite{nie2019dynamic} introduces a lightweight distillator to online distill the pose knowledge via leveraging temporal cues from the previous frame. 
In addition to using local information of adjacent frames, KFP~\cite{zhang2020key} designs a keyframe proposal network that selects informative keyframes after estimating the whole sequence, and then applies a learned dictionary to recover the entire pose sequence. Lastly, MAPN~\cite{fan2021motion} exploits the readily available motion and residual information stored in the compressed streams to dramatically boost the efficiency, and all the residual frames will be calculated by a dynamic gate.

% \vspace{0.1cm}
These proposed methods reduce computation costs by employing adaptive operations on different frames, \ie, complex operations on indispensable frames and simple ones on the rest. 
Despite obtaining efficiency improvement, they still fail to push the efficiency to a higher level since they ignore the fact that it is not necessary to watch each frame. What's more, relying on image features as intermediate representation is heavy for calculation. 

%\vspace{-0.2cm}
\subsection{Motion Completion}
\label{sec:related_completion}
%\vspace{-0.1cm}
Motion completion is widely explored in the area of computer graphics, generally including motion capture data completion~\cite{howarth2010quantitative,reda2018mocap,lai2011motion,gloersen2016predicting,wu2011real,burke2016estimating,kucherenko2018neural,ji2020missing,skurowski2021gap} and motion in-filling~\cite{fragkiadaki2015recurrent,harvey2018recurrent,harvey2020robust,harvey2018recurrent,hernandez2019human,kaufmann2020convolutional,cai2021unified,duan2021single,yuan2021glamr}, which has great significance in the film, animation, and game applications. 
%In terms of different input constraints and various fidelity and diversity requirements, motion in-filling can be further divided into motion prediction, completion, interpolation, and spatial-temporal recovery~\cite{cai2021unified}. 
To be specific, points or sequences missing often occur in motion capture due to technical limitations and occlusions of markers. Accordingly, existing approaches include traditional methods (e.g. linear, Cubic Spline, Lagrange, and Newton's polynomial interpolation, low-rank matrix completion)~\cite{howarth2010quantitative,reda2018mocap,lai2011motion,gloersen2016predicting} and learning-based methods (e.g., Recurrent Neural Networks (RNNs))~\cite{kucherenko2018neural,ji2020missing}. 
Motion in-filling aims to complete the absent poses with specific keyframe constraints. RNNs~\cite{fragkiadaki2015recurrent,harvey2018recurrent,harvey2020robust,harvey2018recurrent,xu2021exploring} and convolutional models~\cite{yan2019convolutional,kaufmann2020convolutional} are commonly used in motion in-filling. 
Recently, Generative adversarial learning~\cite{hernandez2019human,karras2019style} and autoencoder~\cite{kaufmann2020convolutional,cai2021unified} are also introduced for realistic and naturalistic output. Some recent works\cite{duan2021single,ho2021render} also introduce self-attention models to infill the invisible frames.

Although both general motion completion and our target are to recover the full pose sequence, there are two main differences.
On the one hand, the objective of motion completion is to generate diverse or realistic motions under certain assumptions, e.g., a recurring or repeated motion like walking. They may fail when motions are aperiodic and complex. In contrast, our goal is to achieve high efficiency in video-based pose estimation, where the benchmarks are usually from real-life videos.
On the other hand, motion completion assumes having ground-truth poses as inputs rather than estimated poses. Current designs may not be able to handle unreliable and noisy poses generated from deep models.

\section{Method}\label{method}
\label{sec:method}

\subsection{Problem Definition and Overview}
\label{sec: problem definition and motivation}
%\label{sec:method_define}

Given an input video $\cI=\{\mI^t\}{_{t=1}^{T}}$ of length $T$, a pose estimation framework computes the corresponding sequence of poses $\hat{\cP} = \{\hat{\mP}^t\}{_{t=1}^{T}}$, aiming to minimize the distance between the estimated poses $\hat{\cP}$ and the ground-truth poses $\cP$.
$\hat{\mP}^t$ could be any human pose representation, including 2D keypoint position, 3D keypoint position, and 6D rotation matrix.

The main target of this work is to set a baseline for efficient video-based pose estimation without compromising accuracy.
As shown in Fig.~\ref{fig:three_framewors}(c), we devise a three-step \emph{sample-denoise-recover} flow to process video-based pose estimation efficiently and effectively.
As adjacent frames usually contain redundant information and human motion is continuous, \name first samples a small percentage of frames (e.g., $10\%$) $\mathbf{I}^{sampled}$ and applies existing pose estimators~\cite{xiao2018simple,martinez2017simple,joo2020eft,kolotouros2019spin,kocabas2021pare} thereon to obtain the corresponding poses.
However, recovering the full pose sequence from sparsely observed poses is challenging, especially when the poses are estimated by networks and often contain noise.
Relying on a few poses to recover the entire sequence, the quality of sampled poses is the key.
To tackle the challenge, we introduce two subnets, \emph{DenoiseNet} and \emph{RecoverNet}.
%, to form a two-stage procedure.
%
Specifically, \emph{DenoiseNet} refines sparse poses from pose estimator.
Then \emph{RecoverNet} performs motion recovery based on the refined sparse poses to recover the whole pose sequence, with the intuition that humans can perceive complete motion information through a small number of keyframes.
With this new mechanism, the computation cost can be reduced significantly by watching only a small number of frames, which replaces high-cost image feature extraction and pose estimation with a low-cost pose recovery.

\begin{figure}[ht]	
\centering
 	{
 		\begin{minipage}[t]{1.0\linewidth}
 			\centering         
 			\includegraphics[width=4.5in]{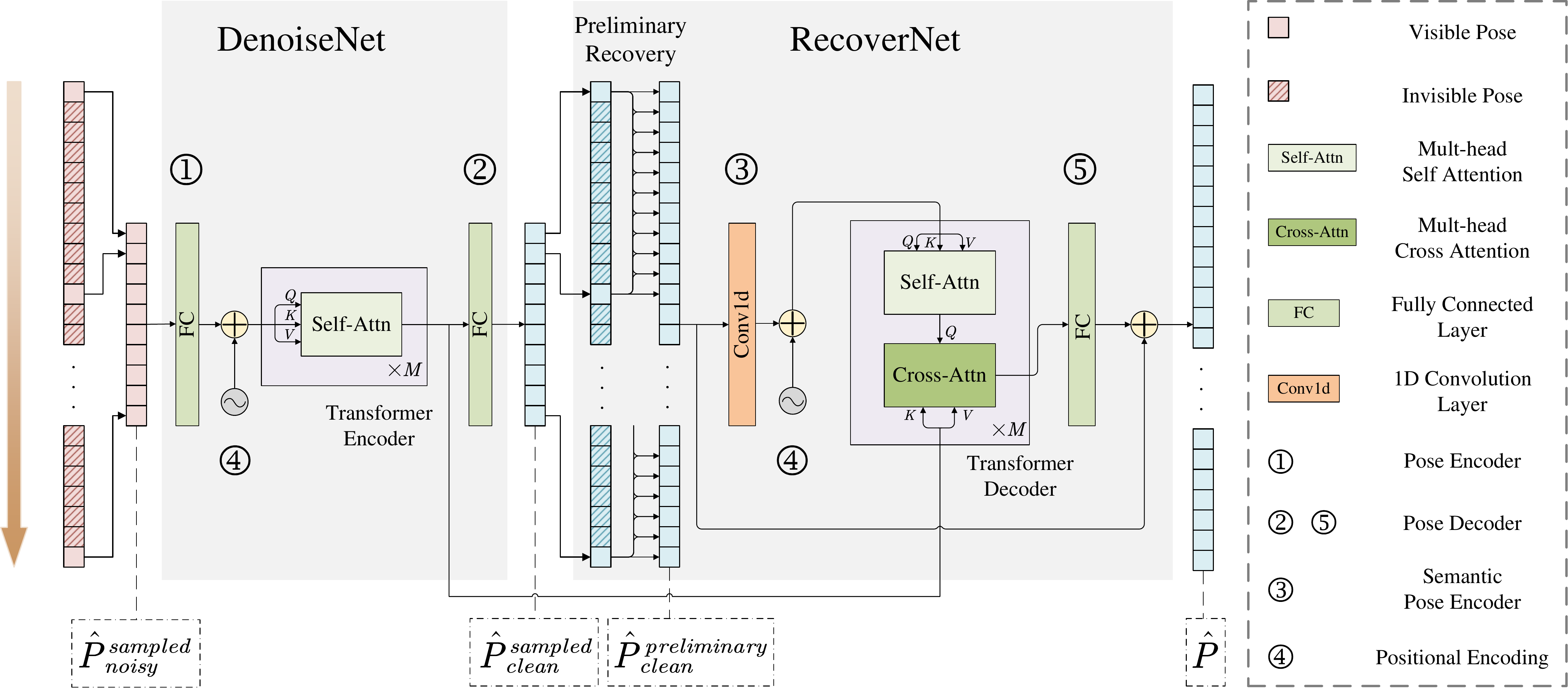}   
 		\end{minipage}
 	} 
\caption{Illustration of the \emph{denoise and recovery} subnets. First, we denoise the sparsely sampled poses $\mathbf{\hat{P}}^{sampled}_{noisy}$ into a clean poses $\mathbf{\hat{P}}^{sampled}_{clean}$ by a transformer-based \emph{DenoiseNet} to handle the dynamically various noises. Then, after a preliminary pose recovery, we embed the sequence into temporal semantic pose tokens and put them into another transformer-based \emph{RecoverNet} that can leverage spatio-temporal correlations to recover realistic and natural poses.}
\label{fig:our_framework} 
% \vspace{-0.4cm}
\end{figure}

%

%\vspace{-0.1cm}
\subsection{Getting Sampled Poses}
%\vspace{-0.1cm}
Different from the previous keyframe-based efficient frameworks~\cite{zhang2020key,fan2021motion} using each frame's feature to select keyframes, we use a uniform sampling that watches one frame in every $N$ frame to select sparse frames $\mathbf{I}^{sampled}$ as a baseline strategy. 
Due to the redundancy in consecutive frames and continuity of human poses, a uniform sampling strategy under a certain ratio is capable of keeping enough information for recovery. 
Then we can estimate $\mathbf{I}^{sampled}$ by any existing pose estimators, such as SimplePose~\cite{xiao2018simple} for 2D poses, FCN~\cite{martinez2017simple} for 3D poses, and PARE~\cite{kocabas2021pare} for 3D body mesh recovery, to get sparse poses $\mathbf{\hat{P}}^{sampled}_{noisy} \in \mathbb{R}^{\frac{T}{N} \times (K \cdot D)}$. $K$ is the number of keypoints, and $D$ is the dimensions for each keypoint. %
Notably, we experimentally show that uniform sampling can surpass complex keyframe selection methods from both efficiency and accuracy in Sec.~\ref{sec:exp_comp_efficient}.

%\vspace{-0.1cm}
\subsection{Denoising the Sampled Poses}
%\vspace{-0.1cm}

Motion completion often resorts to ground-truth sparse poses for infilling the whole sequence. However, in our scenario, the sampled poses are obtained by single-frame pose estimators, inevitably leading to noisy, sparse poses.
Consequently, the quality of sparse poses is crucial for motion recovery. Before recovering the full motion, we develop a denoising network to refine the sampled poses $\mathbf{\hat{P}}^{sampled}_{noisy}$ to clean poses $\mathbf{\hat{P}}^{sampled}_{clean}$.
Due to the temporal sparseness and noisy jitters, the key designs of \emph{DenoiseNet} lie in two aspects: (\romannumeral1) A \emph{dynamic} model for handling diverse possible pose noises; (\romannumeral2) \emph{Global} temporal receptive fields to capture useful Spatio-temporal information while suppressing distracting noises.
Based on these two considerations, local operations, like convolutional or recurrent networks, are not well suited. 
Intuitively, Transformer-based models~\cite{vaswani2017attention} are capable of capturing the global correlations among discrete tokens, so we use Transformer-based encoder modules to relieve noises from the sparse poses. 
The denoise process can be formulated as:
\begin{equation}
\small
\mathbf{\hat{F}}_{clean}^{sampled}=\mathbf{TransformerEncoder}\left( \mathbf{\hat{P}}_{noisy}^{sampled}\mW_{DE}+\mE_{pos} \right)
\label{eq:transformerencoder_1}
\end{equation}
As demonstrated in the left block of Fig.~\ref{fig:our_framework}, after being encoded through a linear projection matrix $\mW_{DE}\in \mathbb{R}^{(K\cdot D)\times C}$,  summed with a positional embedding $\mE_{pos}\in \mathbb{R}^{\frac{T}{N} \times C}$, and then processed by the $\mathbf{TransformerEncoder}$ composed of $M$ Multi-head Self-Attention blocks as in~\cite{vaswani2017attention}, input noisy poses $\mathbf{\hat{P}}_{noisy}^{sampled}$ are embedded into a clean feature $\mathbf{\hat{F}}_{clean}^{sampled}\in \mathbb{R}^{\frac{T}{N}\times C}$, where $C$ is the embedding dimension. 
Dropout, Layer Normalization, and Feedforward layers are the same as the original Transformer.  Lastly, we use another linear projection matrix $\mW_{DD}\in \mathbb{R}^{C \times (K\cdot D)}$ to obtain refined sparse poses.
\begin{equation}
\small
\mathbf{\hat{P}}_{clean}^{sampled}=\mathbf{\hat{F}}_{clean}^{sampled}\mW_{DD}
\label{eq:transformerencoder_2}
\end{equation}
The learnable parameters in \emph{DenoiseNet} are trained by minizing $\mathbf{\hat{P}}_{clean}^{sampled}$ with sampled ground-truth poses $\mathbf{P}^{sampled}$.

\subsection{Recovering the Sampled Poses}
After getting the sparse clean poses $\mathbf{\hat{P}}^{sampled}_{clean}\in \mathbb{R}^{\frac{T}{N} \times (K\cdot D)}$, we use another Spatio-temporal subnet, \emph{RecoverNet}, to recover the absent poses. 
In order to learn the consistent temporal correlations, a simple temporal upsampling (e.g., a linear transformation $\mW_{PR}\in \mathbb{R}^{T \times \frac{T}{N}}$) is applied to perform preliminary sequence recovery to get $\hat{\mP}^{preliminary}_{clean}\in \mathbb{R}^{T \times (K\cdot D)}$. 
\begin{equation}
    \mathbf{\hat{P}}^{preliminary}_{clean}=\mW_{PR}\mathbf{\hat{P}}_{clean}^{sampled}
    \label{eq:recover_1}
\end{equation}
To make the recovery more realistic and accurate, we adopt another transformer-based network for detailed poses recovery. Unlike the previous pose transformers~\cite{zheng20213d}, we bring temporal semantics into pose encoding to encode the neighboring $D$ frames' poses into pose tokens via a temporal 1D convolutional layer. 
The main architecture of \emph{RecoverNet} is also the same as Transformer, which employs $M$ multi-head self-attention blocks.
\begin{equation}
\small
\mathbf{\hat{P}}=\textbf{TransformerDecoder}\left(Conv1d\left( \mathbf{\hat{P}}_{clean}^{preliminary} \right) +\mE_{pos}, \mathbf{\hat{F}}_{clean}^{sampled} \right)\mW_{RD}, 
\label{eq:reover_2}
\end{equation}
where the pose decoder is $\mW_{RD}\in \mathbb{R}^{C \times (K\cdot D)}$.
As illustrated in the second block marked as \emph{RecoverNet} in Fig.\ref{fig:our_framework}, we draw key information in the Cross-Attention block by leveraging denoised features $\mathbf{\hat{F}}_{clean}^{sampled}$.

\subsubsection{Efficiency calculation.}
The computational costs of \name is from three parts: (\romannumeral1) using existing backbones to estimate the sampled poses $\mathbf{\hat{P}}^{sampled}_{noisy}$, (\romannumeral2) using \emph{DenoiseNet} to get clean sampled poses $\mathbf{\hat{P}}^{sampled}_{clean}$, and (\romannumeral3) using \emph{RecoverNet} to recover the clean sampled poses to the complete pose sequence $\mathbf{\hat{P}}^t$. 
To summarize, FLOPs of \name is:

\begin{equation}
\small
    FLOPs = \frac{1}{T}(T/N * f(E) + T*(f(D)+f(R))),
    \label{eq:flop}
\end{equation}
where $f(\cdot)$ calculates the model's per frame FLOPs.
$f(E)$, $f(D)$ and $f(R)$ represent per frame FLOPs of pose estimators, \emph{DenoiseNet} and \emph{RecoverNet}, respectively. Using poses instead of image features as representation makes two subnets computational efficient. Notably, $(f(D)+f(R)) \ll f(E)$ (more than $10^4\times$. Details can be find in Table~\ref{tab:penn_jhmdb} and \ref{tab:3dpose}). Since \name samples very few frames in step $1$, the mean FLOPs can be reduced to $1/N$ compared with those watch-every-frame methods, resulting in a $10\times$ speedup overalls.

\subsection{Loss Function} 
We follow recent 3D pose estimation methods~\cite{pavllo20193d,zeng2021learning} to apply a simple L1 regression loss to minimize the errors between $\mP^t$ and $\hat{\mP}^t$ for 2D or 3D pose estimation. Particularly, to learn the noisy patterns from sampled estimated poses, we further add an L1 loss between sparse estimated poses $\hat{\mP}_{clean}^{sampled}$ and the corresponding ground-truth poses $\mP^{sampled}$. Therefore, the objective function is defined as follows.
\begin{equation}
\small
    \mathcal{L} = \lambda(\frac{1}{T}\sum_{t=1}^{T}|\hat{\mP}^t - \mP^t|) + \frac{1}{(T/N)}\sum_{n=1}^{T/N}|\hat{\mP}_{clean}^{sampled(n)} - \mP^{sampled(n)}|,
    \label{eq:loss}
\end{equation}
%sparse sample n
where $\lambda$ is a scalar to balance the losses between \emph{RecoverNet} and \emph{DenoiseNet}. We set $\lambda=5$ by default.
\section{Experiments\protect\footnote{Due to the limit of pages, we present \emph{data description, comprehensive results of different sampling ratios, the effect of hyper-parameters, generalization ability, qualitative results, and failure cases analyses} in the supplementary material.}}
\label{experiment}

\subsection{Experimental Settings} 
\label{sec:exp_setting} 
%\vspace{5pt}
\noindent \textbf{Datasets.}
We verify our baseline framework on three tasks. For 2D pose estimation, we follow existing video-based efficient methods~\cite{zhang2020key,fan2021motion} using dataset Sub-JHMDB~\cite{jhuang2013towards}. For 3D pose estimation, we choose the most commonly used dataset Human3.6M~\cite{ionescu2013human3}. For 3D body recovery, we evaluate on an in-the-wild dataset 3DPW \cite{von2018recovering} and a dance dataset AIST++ \cite{li2021aist} with fast-moving and diverse actions.

%\vspace{5pt}
\noindent \textbf{Evaluation metrics.}
For 2D pose estimation, we follow previous works~\cite{nie2019dynamic,zhang2020key,fan2021motion} adopting the Percentage of the Correct Keypoints (\emph{PCK}), where the matching threshold is set as 20\% of the bounding box size under pixel level. For 3D pose estimation and body recovery, following~\cite{kanazawa2018hmr,kolotouros2019spin,kocabas2021pare,martinez2017simple,zeng2021smoothnet}, we report Mean Per
Joint Position Error (\emph{MPJPE}) and the mean Acceleration error (\emph{Accel}) to respectively measure the localization precision and smoothness.
Besides, we report efficiency metrics mean FLOPs (G) per frame, the number of parameters and the inference time tested on a single TITAN Xp GPU.

%\vspace{5pt}
\noindent \textbf{Implementation details.}
To facilitate the training and testing in steps $2$ and $3$, we first prepare the detected poses on both training and test sets offline. The uniform sampling ratio is set to $10\%$ by default, which means watching one frame in every $N=10$ frames in videos. To deal with different input video lengths, we input non-overlapping sliced windows with fixed window sizes. It is important to make sure the first and last frames are visible, so the input and output window sizes are both $(N*Q+1)$, where $Q$ is the average number of visible frames in a window. We set $Q=1$ for 2D poses due to the short video length of the 2D dataset and $Q=10$ for others. We change embedding dimension $C$ and video length $T$ to adapt different datasets and estimators, which influence FLOPs slightly.
For \emph{DenoiseNet}, we apply $M=5$ transformer blocks with embedding dimension $C=128$ by default. 
For \emph{RecoverNet}, we use the same settings as \emph{DenoiseNet}. The temporal kernel size of the semantic pose encoder is $5$. For more details, please refer to the supplementary material.
All experiments can be conducted on a single TITAN Xp GPU.

%\vspace{-0.3cm}
\subsection{Comparison with Efficient Video-based Methods}
\label{sec:exp_comp_efficient}
%\vspace{-0.1cm}
Existing efficient video-based pose estimation methods~\cite{nie2019dynamic,zhang2020key,fan2021motion} only validate on 2D poses.
In this section, we compare the accuracy and the efficiency of \name with SOTAs. We follow their experiment settings for fair comparisons and use the same pose estimator SimplePose~\cite{xiao2018simple}.

\begin{table}[ht]
\small
\centering
%\vspace{-0.3cm}
\caption{\textbf{Comparison on Sub-JHMDB~\cite{jhuang2013towards} dataset with existing video-based efficient methods~\cite{nie2019dynamic,zhang2020key,fan2021motion} for 2D pose estimation.} \emph{R} stands for ResNet backbone~\cite{he2016deep}. \emph{Ratio} represents the sampling ratio. The pose estimator of \name is the single-frame model SimplePose (R50)~\cite{xiao2018simple}. Best results are in bold.}
\label{tab:penn_jhmdb}
\setlength{\tabcolsep}{0.5mm}{%
\begin{tabular}{lccccccc|ccc}
\toprule
\multicolumn{11}{l}{\cellcolor{Gray}\textbf{Sub-JHMDB dataset - 2D Pose Estimation}  }                                                                                                  \\ \midrule[0.25pt]
Methods    & Head     & Sho.     & Elb. & Wri. & Hip  & Knee & Ank. & Avg. $\uparrow$& FLOPs(G) $\downarrow$& \emph{Ratio} \\ \midrule
Luo \etal~\cite{luo2018lstm}        & 98.2 & 96.5 & 89.6 & 86.0 & 98.7 & 95.6 & 90.0 & 93.6 & 70.98    & 100\%         \\
DKD (R50)~\cite{nie2019dynamic}        & 98.3 & 96.6 & 90.4 & 87.1 & 99.1 & 96.0 & 92.9 & 94.0 & 8.65     & 100\%        \\
KFP (R50)~\cite{zhang2020key}        & 95.1 & 96.4 & 95.3 & 91.3 & 96.3 & 95.6 & 92.6 & 94.7 & 10.69$^{\dagger}$ & 44.5\% \\
KFP (R18)~\cite{zhang2020key} & 94.7 & 96.3 & 95.2 & 90.2 & 96.4 & 95.5 & 93.2 & 94.5 & 7.19$^{\dagger}$ & 40.8\% \\
MAPN (R18)~\cite{fan2021motion} & 98.2 & 97.4 & 91.7 & 85.2 & 99.2 & 96.7 & 92.2 & 94.7 & 2.70 & 35.2\% \\\hline
SimplePose~\cite{xiao2018simple}  & 97.5 & 97.8 & 91.1 & 86.0 & 99.6 & 96.8 & 92.6 & 94.4 & 11.96    &100\%          \\ 
\textbf{DeciWatch} &  99.8 &99.5& \textbf{99.7}  & \textbf{99.7}  &\textbf{98.7} &99.4& 96.5 & 98.8 & 1.196+0.0005$^{\ddagger}$  &   10.0\%       \\
\textbf{DeciWatch} &  \textbf{99.9} &\textbf{99.8}& 99.6  & \textbf{99.7}  &98.6 &\textbf{99.6}& \textbf{96.6} &  \textbf{98.9} &  0.997+0.0005$^{\ddagger}$ &   8.3\%\\
\textbf{DeciWatch} &  98.4 &98.3& 98.2  & 98.7  & 97.5 &98.3& 95.2 & 97.5 & \textbf{0.598+0.0005}$^{\ddagger}$  &    \textbf{5.0\%}\\ \toprule
\end{tabular}%
}
\begin{tablenotes} 
\tiny
		\item  $^{\dagger}$~The results are recalculated according to \emph{Ratio} and their tested FLOPs for SimplePose (i.e., 11.96G). \\
		\item  $^{\ddagger}$~Tested with ptflops v0.5.2~\cite{SovrasovFlopsCounterConvolutional2022}.
 \end{tablenotes}
% \vspace{-0.5cm}
\end{table}

As shown in Table~\ref{tab:penn_jhmdb}, our approach shows significantly increased accuracy with the highest efficiency, achieving more than $20\times$ improvement in the computation cost on the Sub-JHMDB dataset.
Compared to the SOTA method~\cite{fan2021motion}, we surpass them by $4.3\%$ and $4.4\%$ on average \emph{PCK} (Avg.) with $55.7\%$ and $77.9\%$ reduction in FLOPs. 
Our improvement mainly comes from elbows (Elb.) (from 91.7\% to 99.6\%) and ankles (Ank.) (from 92.2\% to 96.6\%) under a $8.3\%$ ratio. These outer joints usually move faster than inner joints (e.g., Hips), which may cause motion blur and make estimators hard to detect precisely.
However, previous efficient video-based pose estimation methods did not consider a denoising or refinement strategy. \name uses \emph{DenoiseNet} to reduce noises. Then, \emph{RecoverNet} interpolates the sparse poses using the assumption of continuity of motion without watching blurry frames, showing the superiority of \name.

To further verify the effectiveness of the denoise scheme in \name, we input the full sequence of outputs from SimplePose, which means the \emph{Ratio} is $100\%$, and the result of $PCK$ is $99.3$. The additional improvement in accuracy shows that \name can also be used as an effective denoise/refinement model to further calibrate the output positions. 
Based on the above observations, using a lightweight \name in a regression manner to further refine heatmap-based 2D pose estimation methods can be a promising refinement strategy.
For efficiency, the total number of parameters in \emph{DenoiseNet} and \emph{RecoverNet} is $0.60$M and the inference time is about $0.58$ms/frame.

Besides, we argue that PCK@0.05 with lower thresholds will be better to reflect the effectiveness of the methods since the commonly used metric PCK@0.2 appears saturated. We report PCK@0.05 and PCK@0.1 of \name in Supp.

\subsection{Boosting Single-frame Methods}
\label{sec:exp_comp_single}

\noindent \textbf{The used single-frame pose estimators:} We compare \name with the following single-frame pose estimators~\cite{martinez2017simple,kolotouros2019spin,joo2020eft,kocabas2021pare} that watch each frame when estimating a video. 
We first introduce these methods as follows.

\noindent -- FCN~\cite{martinez2017simple} is one of the most important 2D-to-3D methods with multiple fully connected layers along the spatial dimension.

\noindent -- SPIN~\cite{kolotouros2019spin} is one of the most commonly used methods, which combines SMPL optimization in the training process.

\noindent -- EFT~\cite{joo2020eft} is trained on augmented data compared with SPIN~\cite{kolotouros2019spin} to get better performance and generalization ability.

\noindent -- PARE~\cite{kocabas2021pare} proposes a part-guided attention mechanism to handle partial occlusion scenes, achieving the state-of-the-art on many benchmarks.

\begin{table}[h]
\centering
\small
%\vspace{-0.3cm}
\caption{\textbf{Comparing \name with existing single-image 3D pose estimators on Human3.6M~\cite{ionescu2013human3}, 3DPW~\cite{von2018recovering}, and AIST++~\cite{li2021aist} datasets.} Pose estimators used in \name keep the same as the corresponding methods.}
\label{tab:3dpose}
\setlength{\tabcolsep}{1mm}{%
\begin{tabular}{l|c|c|c|c}
\toprule
Methods     & \multicolumn{1}{c}{\emph{MPJPE} $\downarrow$}& \multicolumn{1}{c}{\emph{Accel} $\downarrow$}& FLOPs(G)$\downarrow$&  \emph{Ratio} \\
\midrule
\multicolumn{5}{l}{\cellcolor{Gray}\textbf{Human3.6M~\cite{ionescu2013human3} - 3D Pose Estimation}}
\\   \midrule[0.25pt]
FCN~\cite{martinez2017simple}  &54.6&19.2&6.21& 100.0\% \\
\textbf{DeciWatch} &  \textbf{52.8}{\color{Red}$\downarrow_{1.8(3.3\%)}$} &1.5{\color{Red}$\downarrow_{17.7(92.2\%)}$}&0.621+0.0007&  \textbf{10.0\%} \\
\textbf{DeciWatch} & 53.5{\color{Red}$\downarrow_{1.1(2.0\%)}$} &\textbf{1.4}{\color{Red}$\downarrow_{17.8(92.7\%)}$}&0.414+0.0007& \textbf{6.7\%} \\\toprule
 
\multicolumn{5}{l}{\cellcolor{Gray}\textbf{3DPW~\cite{von2018recovering} - 3D Body Recovery} }    
                                    \\ \midrule[0.25pt]
SPIN~\cite{kolotouros2019spin}  &96.9&34.7&4.13& 100.0\% \\
\textbf{DeciWatch} & \textbf{93.3}{\color{Red}$\downarrow_{3.6(3.7\%)}$} &7.1{\color{Red}$\downarrow_{27.6(79.5\%)}$}&0.413+0.0004&  \textbf{10.0\%} \\
\textbf{DeciWatch} &  96.7{\color{Red}$\downarrow_{0.2(0.2\%)}$}&\textbf{6.9}{\color{Red}$\downarrow_{27.8(80.1\%)}$}&0.275+0.0004& \textbf{6.7\%} \\\hline
EFT~\cite{joo2020eft}  &90.3&32.8&4.13&  \textbf{100.0\%} \\
\textbf{DeciWatch} &\textbf{89.0}{\color{Red}$\downarrow_{1.3(1.4\%)}$}&6.8{\color{Red}$\downarrow_{26.0(79.3\%)}$}&0.413+0.0004& \textbf{10.0\%} \\
\textbf{DeciWatch} & 92.3{\color{Blue}$\uparrow_{2.0(2.2\%)}$} &\textbf{6.6}{\color{Red}$\downarrow_{26.2(79.9\%)}$}&0.275+0.0004&  \textbf{6.7\%} \\\hline
PARE~\cite{kocabas2021pare}  &78.9&25.7&15.51& 100.0\% \\
\textbf{DeciWatch} &\textbf{77.2}{\color{Red}$\downarrow_{1.7(2.2\%)}$}&6.9{\color{Red}$\downarrow_{18.8(73.2\%)}$}&1.551+0.0004& \textbf{10.0\%} \\
\textbf{DeciWatch} & 80.7{\color{Blue}$\uparrow_{1.8(2.2\%)}$} &\textbf{6.7}{\color{Red}$\downarrow_{18.6(73.9\%)}$}&1.034+0.0004& \textbf{6.7\%} \\\toprule

\multicolumn{5}{l}{\cellcolor{Gray}\textbf{AIST++~\cite{li2021aist} - 3D Body Recovery}     }
                                    \\ \midrule[0.25pt]
SPIN~\cite{kolotouros2019spin}  &107.7&33.8& 4.13&100.0\% \\
\textbf{DeciWatch} &  \textbf{71.3}{\color{Red}$\downarrow_{36.4(33.8\%)}$}&5.7{\color{Red}$\downarrow_{28.1(83.1\%)}$}&0.413+0.0007&  \textbf{10.0\%} \\
\textbf{DeciWatch} &  82.3{\color{Red}$\downarrow_{25.4(23.6\%)}$}&\textbf{5.5}{\color{Red}$\downarrow_{28.3(83.7\%)}$}&0.275+0.0007& \textbf{6.7\%} \\\bottomrule
\end{tabular}}

\begin{tablenotes} 
\tiny
		\item  All estimation results are re-implemented or tested by us for fair comparisons.
 \end{tablenotes}
%\vspace{-0.5cm}
\end{table}

% \vspace{5pt}
\noindent \textbf{The comparisons:} We demonstrate the comparison results in Table~\ref{tab:3dpose} at sampling ratios of $10\%$ ($N=10$) and $6.7\%$ ($N=15$). To be specific, when the sampling ratio is $10\%$, \name can reduce \emph{MPJPE} by about $2\%$ to $3\%$ for most estimators and reduce \emph{Accel} by about $73\%$ to $92\%$, indicating \name achieves higher precision and smoothness with about $10\%$ FLOPs. Moreover, with $6.7\%$ watched frames, \name still has the capability to recover the complete pose sequence with competitive results.
For the AIST++ dataset, we surprisingly find that training on sparse poses and recovering them can significantly improve output qualities by $33.8\%$ and $23.6\%$ with a sampling ratio of $10\%$ and $6.7\%$ respectively. This indicates that our method is capable of datasets with fast movements and difficult actions, such as Hip-hop or Ballet dances.

In general, we attribute the high efficiency of \name to the use of lightweight and temporal continuous poses representation rather than the heavy features used by previous works~\cite{nie2019dynamic,zhang2020key,fan2021motion}. Meanwhile, the superior effectiveness, especially for motion smoothness, comes from its ability to capture spatio-temporal dynamic relations in the denoising and recovery process and the well-designed sample-denoise-recover steps. Additionally, the inference speeds in step $2$ and $3$ are about $0.1$ms/frame, significantly faster than image feature extraction.

%\vspace{-0.3cm}
\subsection{Comparison with Motion Completion Techniques}
\label{sec:exp_comp_interp}

The third step of \name is similar to motion completion/interpolation as introduced in Sec.~\ref{sec:related_completion}.
To assess existing interpolation methods quantitatively, we compare our model with four traditional methods and one of the latest learning-based interpolation methods~\cite{cai2021unified} based on Conditional Variational Auto-Encoder(CVAE). 
The original experiments in the CAVE-based model are based on the ground-truth of the Human3.6M dataset~\cite{ionescu2013human3} (marked as CVAE~\cite{cai2021unified}-R.$\star$). We compare two additional settings on the same dataset: (\romannumeral1). CVAE~\cite{cai2021unified}-R. inputs estimated 3D poses rather than ground-truth 3D poses and uses Random sampling; (\romannumeral2). CVAE~\cite{cai2021unified}-U. inputs estimated 3D poses and use Uniform sampling, which is the same setting as \name.
For a fair comparison, we adjusted the sampling ratio of training and testing to be consistent as $20\%$, $10\%$, and $5\%$.

\begin{table}[h]
%    \vspace{-0.3cm}
	\centering
	\small
% 	\footnotesize
	\scriptsize
	%\tiny
    \caption{\textbf{Comparison of \emph{MPJPE} with existing motion completion methods on Human3.6M dataset~\cite{ionescu2013human3} for 3D pose estimation.}  Noted that~\cite{cai2021unified} is originally trained and tested on ground-truth 3D poses (noted by $\star$) with random sampling (CVAE~\cite{cai2021unified}-R.), we retrain their model with detected 3D poses to keep the same uniform sampling as us (CVAE~\cite{cai2021unified}-U.). We use FCN~\cite{martinez2017simple} as the single-frame estimator to generate the sparse detected results, and its \emph{MPJPE} is $54.6$mm. }
	{%
		\begin{tabular}{l|cccc|ccc|c}

			\specialrule{.1em}{.05em}{.05em}
			\emph{Ratio}&Nearest& Linear&Quadratic & Cubic-Spline&CVAE-R.$\star$&CVAE-R.& CVAE-U. & \name\\
			\midrule
		    20\%&54.4&54.4&\underline{54.3}&54.5&87.4&114.1&119.4& \textbf{52.8}{\color{Red}$\downarrow_{1.8(3.2\%)}$}\\
		    10\%&54.7&\underline{54.3}&55.2&54.4&99.1&119.2&121.5&\textbf{52.8}{\color{Red}$\downarrow_{1.8(3.2\%)}$} \\
		    5\%&57.6&57.5&\underline{57.2}&57.3&134.9& 140.5 &123.1&\textbf{54.4}{\color{Red}$\downarrow_{0.2(0.3\%)}$}  \\
		    
        \midrule
        \end{tabular}%
	}
	\begin{tablenotes} 
    \tiny
		\item  All estimation results are re-implemented or tested by us for fair comparisons.
 \end{tablenotes}
\label{tab:learn_interp}
%\vspace{-0.5cm}
\end{table}

In Table~\ref{tab:learn_interp}, \name outperforms all methods. Specifically, we find that the results of the CVAE-based model are even twice as bad as the traditional methods at all ratios, especially with estimated poses inputs and uniform sampling. This is because CVAE-based methods try to encode a long sequence of motion into an embedding and then recover them, which is practically difficult to embed well and recover precisely for a specific video. 
Instead, our method and the traditional interpolation strategies directly utilize the continuity of human poses as a priori, making the interpolation process easier.

Owing to the relatively low \emph{MPJPE} and the slow motions in the Human3.6M dataset, \name only have limited improvement over traditional methods. We evaluate on a more challenging dance dataset AIST++~\cite{li2021aist} in Table~\ref{tab:traditional_interp}. A tremendous lift is revealed to over $30\%$ under a $10\%$ ratio. The improvement of \name is from: \name can learn to minimize errors with data-driven training, especially poses with high errors, while traditional methods have no such prior knowledge to decrease the errors from both visible and invisible poses.

\begin{table}[h]
%\vspace{-0.3cm}
	\centering
% 	\footnotesize
	% \scriptsize
	%\tiny
    \caption{\textbf{Comparison of \emph{MPJPE} with traditional interpolation methods on AIST++ dataset~\cite{li2021aist}.} We use 3D pose estimator SPIN~\cite{kolotouros2019spin} as the single-frame estimator, and its \emph{MPJPE} is $107.7$mm.}
	{%
		\begin{tabular}{l|cccc|c}

			\specialrule{.1em}{.05em}{.05em}
			
			\emph{Ratio}&Nearest& Linear&Quadratic & Cubic-Spline& \name\\
			\midrule
		    20\%&106.7 & \underline{104.6}& 105.8& 106.8&\textbf{67.6}{\color{Red}$\downarrow_{39.7(37.0\%)}$}\\
		    10\%&108.3 &\underline{106.3} &108.2 & 108.9&\textbf{71.3}{\color{Red}$\downarrow_{36.0(33.6\%)}$}\\
		    5\%&123.2 & 120.7& \underline{119.9} &121.2 &\textbf{90.8}{\color{Red}$\downarrow_{16.5(15.4\%)}$}\\
        \midrule
        \end{tabular}%
	}
	\label{tab:traditional_interp}
%\vspace{-0.5cm}
\end{table}

%\vspace{-0.3cm}
\subsection{Ablation Study}
\label{sec:exp_ablation}
%\vspace{-0.2cm}
As a baseline framework, we do not emphasize the novelty of network design but provide some possible designs in each step for further research. We have explored the impact of different pose estimators in Table~\ref{tab:penn_jhmdb} and \ref{tab:3dpose} in previous sections. This section will explore how designs in steps $2$ and $3$ influence the results. All experiments use the same input window length at $101$ and a $10\%$ sampling ratio by default. We keep the same setting in both training and testing.

\noindent \textbf{Impact of sampling ratio and input window size.}
%W=t*n+1, long coverage
Due to space limitation, we discuss this part \emph{in Supp.}. We summarize the key observations as follows :(\romannumeral1). With the increase in sampling ratio, the \emph{MPJPEs} first drop before rising, and they are at the lowest when the sampling ratio is about 20\%. \emph{Accels} will decrease constantly. These observations give us a new perspective that in pose estimation, \emph{it is not essential to watch all frames for achieving a better and smoother performance}. (\romannumeral2). Besides, the \emph{MPJPE} of \name surpasses the original baseline even when the sampling ratio is about $8\%$. (\romannumeral3). Lastly, \name is robust to different window sizes from $11$ to $201$ frames.

\setlength{\tabcolsep}{2pt}
\begin{wraptable}{r}{0.5\textwidth}
%\begin{table}[h]
%    \vspace{-1.2cm}
	\centering
	\small
% 	\scriptsize
    \caption{\textbf{Comparison of \emph{MPJPE} with different sampling strategies on 3DPW dataset with EFT~\cite{joo2020eft} pose estimator (\emph{MPJPE} is $90.3$mm).} \emph{U.-2} and \emph{U.-3} are uniform sampling $2$ or $3$ frames for every $N$ frames. \emph{U.-R.} conducts both uniform sampling and random sampling. \emph{R.} is random sampling.}
%    \vspace{-1pt}
	{%
		\begin{tabular}{l|c|c|c|c|c}

			\specialrule{.1em}{.05em}{.05em}
			
			\emph{Ratio}&\emph{U.(Ours)}&\emph{U.-2}&\emph{U.-3}&\emph{U.-R.}&\emph{R.} \\
			\midrule
		    20\% &\textbf{87.2}&89.3&91.5&94.2&97.1\\
		    10\% &\textbf{89.0}&96.3&103.4&101.3&104.4\\
        \midrule
        \end{tabular}%
	}
	\label{tab:sample_strategies}
%\end{table}
%\vspace{-1cm}
\end{wraptable}

\noindent \textbf{Impact of sampling strategies.}
Although we use uniform sampling one frame for every N frame, there are other sampling strategies that can be adopted without watching each frame, such as (\romannumeral1). uniform sampling $2$ or $3$ frames for every N frames, which contains velocity and acceleration information (named as \emph{U.-2} and \emph{U.-3}); (\romannumeral2). random sampling (\emph{R.}); (\romannumeral3). combining uniform sampling with random sampling (\emph{U.-R.}). 
%MPJPEs are reported with 20\% and 10\% sampling ratio.
From Table~\ref{tab:sample_strategies}, \emph{U.-2} and \emph{U.-3} get the worse results compared to \emph{U.(Ours)} because intervals between visible frames become longer, and the information in two or three adjacent frames is too similar to be helpful for the recovery. Moreover, random sampling shows is capable of recovery since a long invisible period may appear, which is hard for model learning. Combining uniform sampling (half of the frames) to avoid long invisible periods can slightly decrease the error in random sampling. In summary, uniform sampling one frame for every N frame (\emph{U.(Ours)}) surpasses all other sampling strategies under the same model.

%\vspace{5pt}
\noindent \textbf{Impact of denoise and recovery subnets.}
%spatial-temporal
In Table~\ref{tab:denoise}, we comprehensively verify the effectiveness of \emph{DenoiseNet} and \emph{RecoverNet} at 10\% sampling ratio on three datasets.
When we remove any part of the two subnets, the results deteriorate to various degrees. Removing \emph{RecoverNet} means we only use a preliminary recovery via a temporal linear layer, which leads to unsatisfying results as discussed in Sec.~\ref{sec:exp_comp_interp}. In fact, \emph{RecoverNet} is very important for the whole framework since it is supervised by the entire sequence's ground-truth, especially for the fast-moving dance dataset AIST++~\cite{li2021aist}. \emph{DenoiseNet} can remove noises in advance while giving a better initial pose sequence to \emph{RecoverNet}, which can reduce the burden in the recovery stage. In summary, the two subnets are both essential and effectively improve the final performance.

\begin{table}[h]
%    \vspace{-0.3cm}
    \small
	\centering
    \caption{\textbf{Exploring impacts of the two \emph{DenoiseNet} and \emph{RecoverNet} subnets with 10\% sampling ratio on three dataset and the corresponding backbones.} $Ori.$ means the original estimator (watching all frames) with $100\%$ sampling ratio. No \emph{RecoverNet} is preliminarily recovered via a temporal linear layer.}
	{%
		\begin{tabular}{l|c|c|c|c}

			\specialrule{.1em}{.05em}{.05em}
			
			Dataset w/Backbone&Ori. ($100$\%)& No \emph{DenoiseNet} & No \emph{RecoverNet} & \name\\
			\midrule
			Human3.6M w/FCN~\cite{martinez2017simple}&54.6&54.5&54.7&\textbf{52.8}\\
		    3DPW w/PARE~\cite{kocabas2021pare}&78.9&79.8&81.0&\textbf{77.2}\\
		    AIST++ w/SPIN~\cite{joo2020eft}&107.7&91.5&95.3&\textbf{71.3}\\
        \midrule
        \end{tabular}%
	}
	\label{tab:denoise}
%\vspace{-0.5cm}
\end{table}

% \vspace{-0.2cm}
\subsection{An application: Efficient Pose Labeling in Videos}
\label{sec:app}
% \vspace{-0.2cm}
% interpolate with sparse ground truth
A large amount of labeled data leads to the success of deep models. However, labeling each frame in videos is labor-intensive and high cost. It is also hard to guarantee continuity among adjacent frames, especially for 3D annotations. Due to the efficiency and smoothness of the pose sequences recovered by \name, reducing the need for dense labeling could be a potential application. We verify the effectiveness of this application on the Human3.6M and AIST++ dataset by directly inputting the sparse ground-truth 3D positions into the \emph{RecoverNet} of \name. Due to limited pages, please see Supp. for details.

% \vspace{-0.2cm}
\section{Conclusion and Future Work}\label{conclusion}
% \vspace{-0.2cm}
This work proposes a sample-denoise-recover flow as a simple baseline framework for highly efficient video-based 2D/3D pose estimation. Thanks to the lightweight representation and continuity characteristics of human poses, this method can watch one frame in every $10$ frame and achieve nearly $10\times$ improvement in efficiency while maintaining competitive performance, as validated in the comprehensive experiments across various video-based human pose estimation and body mesh recovery tasks. There are many opportunities to further improve the proposed baseline solution:

%\vspace{5pt}
\noindent \textbf{Adaptive sampling and dynamic recovery.}  
In \name, we use a simple uniform sampling strategy for all the joints.
In practice, the movements of different joints under different actions vary greatly. Consequently, an adaptive sampling strategy has the potential to further boost the efficiency of video-based pose estimators. For instance, combining multi-modality information (e.g., WIFI, sensors) to relieve visual computation can be interesting. Correspondingly, how to design a dynamic recovery network that can handle non-uniformly sampled poses is an interesting yet challenging problem to explore.

%\vspace{5pt}
\noindent \textbf{High-performance pose estimator design.} 
While this work emphasizes the efficiency of pose estimators, our results show that watching fewer frames with our framework could achieve better per-frame precision compared with watching each frame.
This is in line with the recent findings on multi-view pose estimation methods~\cite{Chu_2021_CVPR,gundavarapu2019structured,shuai2021adaptively}, showing better results without calculating every possible view simultaneously. We attribute the above phenomena to the same intrinsic principle that it is likely to achieve better results by discarding some untrustworthy estimation results. Therefore, designing such a strategy to achieve the best pose estimation performance is an interesting problem to explore. 

\noindent \textbf{Acknowledgement.} This work is supported in part by Shenzhen-Hong Kong-Macau Science and Technology Program (Category C) of Shenzhen Science Technology and Innovation Commission under Grant No. SGDX2020110309500101, and Shanghai AI Laboratory.

% \clearpage\mbox{}Page \thepage\ of the manuscript.
% \clearpage\mbox{}Page \thepage\ of the manuscript.

% This is the last page of the manuscript.
% Now we have reached the maximum size of the ECCV 2022 submission (excluding references).
% References should start immediately after the main text, but can continue on p.15 if needed.

\clearpage

\bibliographystyle{splncs04}

\end{document}

% \linenumbers
\pagestyle{headings}
\mainmatter
\def\ECCVSubNumber{1845}  % Insert your submission number here

\title{---Supplementary Materials---\\DeciWatch: A Simple Baseline for $10\times$ Efficient 2D and 3D Pose Estimation} % Replace with your title

\begin{comment}
\titlerunning{ECCV-22 submission ID \ECCVSubNumber} 
\authorrunning{ECCV-22 submission ID \ECCVSubNumber} 
\author{Anonymous ECCV submission}
\institute{Paper ID \ECCVSubNumber}
\end{comment}
%******************

%\begin{comment}
\titlerunning{DeciWatch}
\author{Ailing Zeng$^{1}$  \and
Xuan Ju$^{1}$  \and
Lei Yang$^{2}$  \and
Ruiyuan Gao$^{1}$  \and
Xizhou Zhu$^{2}$  \and
Bo Dai$^{3}$  \and
Qiang Xu$^{1}$ 
}
\authorrunning{A. Zeng et al.}
\institute{$^{1}$The Chinese University of Hong Kong, $^{2}$Sensetime Group Limited, \\$^{3}$Shanghai AI Laboratory\\
\email{\{alzeng, qxu\}@cse.cuhk.edu.hk}
}
%\end{comment}
%******************
\maketitle

In Sec.~\ref{sec:supp_exp}, we present dataset descriptions. Next, we present results of efficient labeling in Sec.~\ref{sec:supp_app} and the generalization ability of \name in Sec.~\ref{sec:supp_general}. Then, we show more ablation studies on different sampling ratios, model designs of \textit{DenoiseNet} and \textit{RecoverNet}, and hyper-parameters in Sec.~\ref{sec:supp_ablation}. Moreover, we show qualitative comparison results in Sec.~\ref{sec:supp_viz} to demonstrate why \name works. Last, in Sec.~\ref{sec:supp_fail}, we discuss some failure cases in this method to motivate further research.

\section{Dataset Descriptions}
\label{sec:supp_exp}

\noindent -- \textbf{Sub-JHMDB}  JHMDB\cite{jhuang2013towards} is a video-based dataset for 2D human pose estimation. For a fair comparison, we only conduct our experiments on a subset of JHMDB called sub-JHMDB. It contains $316$ videos and the average duration is $35$ frames. For each frame, it provides $15$ annotated body keypoints. We use the bounding box calculated from the puppet mask provided by \cite{luo2018lstm}. Following the settings \cite{zhang2020key,fan2021motion}, we mix $3$ original splitting schemes for training and testing together in 2D pose estimation experiments.

\noindent -- \textbf{Human3.6M}~\cite{ionescu2013human3} Human3.6M is a large-scale indoor video dataset with $15$ actions from $4$ camera viewpoints. It has $3.6$ million frames and a frame rate of $50$ fps. 3D human joint positions are captured accurately from a high-speed motion capture system. Following previous works~\cite{zeng2020srnet,martinez2017simple,pavllo20193d,zeng2021smoothnet}, we use the standard protocol with $5$ actors (S$1$, S$5$, S$6$, S$7$, S$8$) as the training set and another $2$ actors (S$9$, S$11$) as the testing set. 

%We can use the camera intrinsic parameters to calculate their accurate 2D joint positions. 
\noindent -- \textbf{3DPW} \cite{von2018recovering} 3DPW is a challenging in-the-wild dataset consisting of more than $51$k frames with accurate 3D poses and shapes annotation. The sequences are $30$fps. This dataset is usually used to validate the performance of model-based body recovery methods~\cite{kanazawa2018hmr,kolotouros2019spin,joo2020eft,kocabas2021pare}.

\noindent -- \textbf{AIST++} \cite{li2021aist} is a challenging dataset with diverse and fast-moving dances that comes from the AIST Dance Video DB~\cite{tsuchida2019aist}. It contains 3D human motion annotations of $1,408$ video sequences at $60$ fps, which is $10.1$M frames in total. The 3D human keypoint annotations and SMPL parameters it provides cover $30$ different actors in $9$ views. We follow the original settings to split the training and testing sets based on actors and actions.

% \vspace{-0.2cm}
\section{An application: Efficient Pose Labeling in Videos}
\label{sec:supp_app}
% \vspace{-0.2cm}
% interpolate with sparse ground truth
Due to the efficiency and smoothness of the pose sequences recovered by \name, reducing the need for dense labeling could be a potential application. We verify the effectiveness of this application on the Human3.6M and AIST++ dataset by directly inputting the sparse ground-truth 3D positions into the \emph{RecoverNet} of \name. In Table~\ref{tab:label}, we compare \name with the most used spline interpolation, linear interpolation, and quadratic interpolation. Our method has a slower error growth as the interval $N$ gets larger. 
%Marking one frame every 10 frames gives a 2mm error that is acceptable in some application scenarios
%
To be specific, it is possible to label one frame every $10$ frames with only $2.89$mm position errors in slow movement videos (e.g., in Human3.6M~\cite{ionescu2013human3}) and label one frame every $5$ frames with only $4.03$mm position errors in fast-moving videos (e.g., in AIST++~\cite{li2021aist}). This application can improve annotation efficiency by more than $10\times$.
% where its keypoint annotations are captured by Vicon devices to provide precise labels
\begin{table}[h]
%    \vspace{-0.5cm}
    \small
	\centering
    \caption{\textbf{Comparison of \emph{MPJPE} on efficient pose labeling that labels one frame in every $N$ frames on Human3.6M~\cite{ionescu2013human3} and AIST++~\cite{li2021aist} dataset.}}
	{%
		\begin{tabular}{l|ccccc||ccccc}

			\specialrule{.1em}{.05em}{.05em}
			&\multicolumn{5}{c||}{\cellcolor{Gray}\textbf{Human3.6M}}&\multicolumn{5}{c}{\cellcolor{Gray}\textbf{AIST++}}\\
			\cmidrule{2-11}
			Interval N&2 &  5 &10&15 &20&2 & 5 &10&15 &20\\
			\midrule
			Linear&2.21&6.55&10.81&24.15&35.20&7.21&21.31&27.72&73.69&99.04 \\
			Quadratic&1.26&4.31&10.05&\underline{17.22}&\underline{22.85}&2.04&8.33&23.59&\underline{43.13}&\underline{61.16}\\
			Cubic-Spline&\textbf{0.18}&\textbf{0.99}&\underline{5.36}&18.42&29.21&\underline{0.89}&\underline{5.12}&\underline{18.31}&45.32&77.39\\
		    \textbf{\name}& \underline{0.25}&\underline{1.33
		    }&\textbf{2.89}&\textbf{6.21}&\textbf{10.59}&\textbf{0.83}&\textbf{4.03}&\textbf{11.25}&\textbf{20.12}&\textbf{41.25}\\
        \midrule
        \end{tabular}%
	}
	\label{tab:label}
%	\vspace{-1cm}
\end{table}

\section{Additional Evaluation Metrics for 2D Pose Estimation}

As is shown in Tab. 1 in the main paper, the results of \name have achieved nearly 99\% accuracy on PCK@0.2. However, qualitative visualization shows that an awful lot of errors still exist in the recovery results. We attribute it to the fact that PCK@0.2 is quite loose for accuracy measurement, which only requires the detected keypoints to be within 20\% of the bounding box size under pixel level. As a result, we use two additional evaluation metrics, PCK@0.1, and PCK@0.05, for better localization evaluation. More specifically, PCK@0.1 and PCK@0.05 restrict the matching threshold to 10\% and 5\% of the bounding box size. Tab.~\ref{table:pck0050102} shows the results of \name and SimplePose\cite{xiao2018simple} on these three metrics. In future work, we recommend using PCK@0.05 as the main metrics for 2D pose estimation.
 
\begin{table}[H]
\centering
\caption{\textbf{Comparison of \name and SimplePose\cite{xiao2018simple} on PCK@0.2, PCK@0.1, and PCK@0.05}. In future work, we recommend using PCK@0.05 as the main metrics for 2D pose estimation.}
\setlength{\tabcolsep}{2mm}{
\begin{tabular}{c|c|ccc}
\hline
\multicolumn{5}{l}{\cellcolor{Gray}\textbf{Sub-JHMDB dataset - 2D Pose Estimation}}                                         \\ \hline
\multicolumn{1}{l}{Sampling Ratio} & \multicolumn{1}{l}{Evaluation Metric} & PCK@0.2 $\uparrow$          & PCK@0.1 $\uparrow$          & PCK@0.05 $\uparrow$         \\ \hline
                                   & SimplePose                            & 93.92\%          & 81.25\%          & 56.88\%          \\ \cline{2-5} 
\multirow{-2}{*}{20\%}             & \textbf{DeciWatch}                    & \textbf{99.11\%} & \textbf{95.43\%} & \textbf{82.66\%} \\ \hline
                                   & SimplePose                            & 93.94\%          & 81.61\%          & 57.30\%          \\ \cline{2-5} 
\multirow{-2}{*}{10\%}             & \textbf{DeciWatch}                    & \textbf{98.75\%} & \textbf{94.05\%} & \textbf{79.44\%} \\ \hline
                                   & SimplePose                            & 92.38\%          & 82.79\%          & 58.95\%          \\ \cline{2-5} 
\multirow{-2}{*}{5\%}              & \textbf{DeciWatch}                    & \textbf{97.50\%} & \textbf{91.76\%} & \textbf{73.02\%} \\ \hline
\end{tabular}
}
\label{table:pck0050102}

\end{table}

\section{Generalization Ability}
\label{sec:supp_general}
\name learns the patterns of noisy human motions since motion distribution could be overlapped among some datasets, making it has potential generalization ability. 
We further test \name trained on 3DPW-PARE across various backbones and datasets in Tab.~\ref{table:generalize}, where \name still achieves competitive pose estimation results with 10x efficiency. We attribute it to the fact that \name effectively learns the continuity of motions, which is applicable for different sorts of motions.

\begin{table}[H]
	\centering
    \scriptsize
    \caption{\textbf{Cross-backbone and cross-dataset results from \name checkpoints trained on 3DPW-PARE.}}
	{
		\begin{tabular}{l|cccc}
		\midrule[0.25pt]
        Dataset/Estimator            &           & MPJPE$\downarrow$ & Accel.$\downarrow$  \\
        \midrule[0.25pt]
        \multirow{2}{*}{3DPW/EFT}    & Estimator (100\%) & 90.3  & 32.8      \\
                                     &\name (\textbf{10\%})   & \textbf{87.2}{\color{Red}$\downarrow_{3.1(3.4\%)}$}  & \textbf{7.2}{\color{Red}$\downarrow_{25.6(77.9\%)}$}     \\
        \midrule[0.15pt]
        \multirow{2}{*}{3DPW/SPIN}   & Estimator (100\%) & \textbf{96.9}  & 34.6      \\
                                     & \name (\textbf{10\%})   & 98.3{\color{Blue}$\uparrow_{1.4(1.4\%)}$}  & \textbf{7.1}{\color{Red}$\downarrow_{27.5(79.5\%)}$}       \\
        \midrule[0.15pt]
        \multirow{2}{*}{AIST++/SPIN} & Estimator (100\%) & 107.7 & 33.8   \\
                                     & \name (\textbf{10\%})  & \textbf{101.8}{\color{Red}$\downarrow_{5.9(5.5\%)}$} & \textbf{6.2}{\color{Red}$\downarrow_{27.6(81.7\%)}$}   \\
        \midrule[0.25pt]
        \end{tabular}
	}

	\label{table:generalize}
\end{table}

\section{Ablation Study}
\label{sec:supp_ablation}

\begin{figure}[h]	
\centering
    \subfigure[Effect of the sampling ratio] 
	{
		\begin{minipage}[t]{0.46\linewidth}
			\centering      
			\includegraphics[width=2.3in]{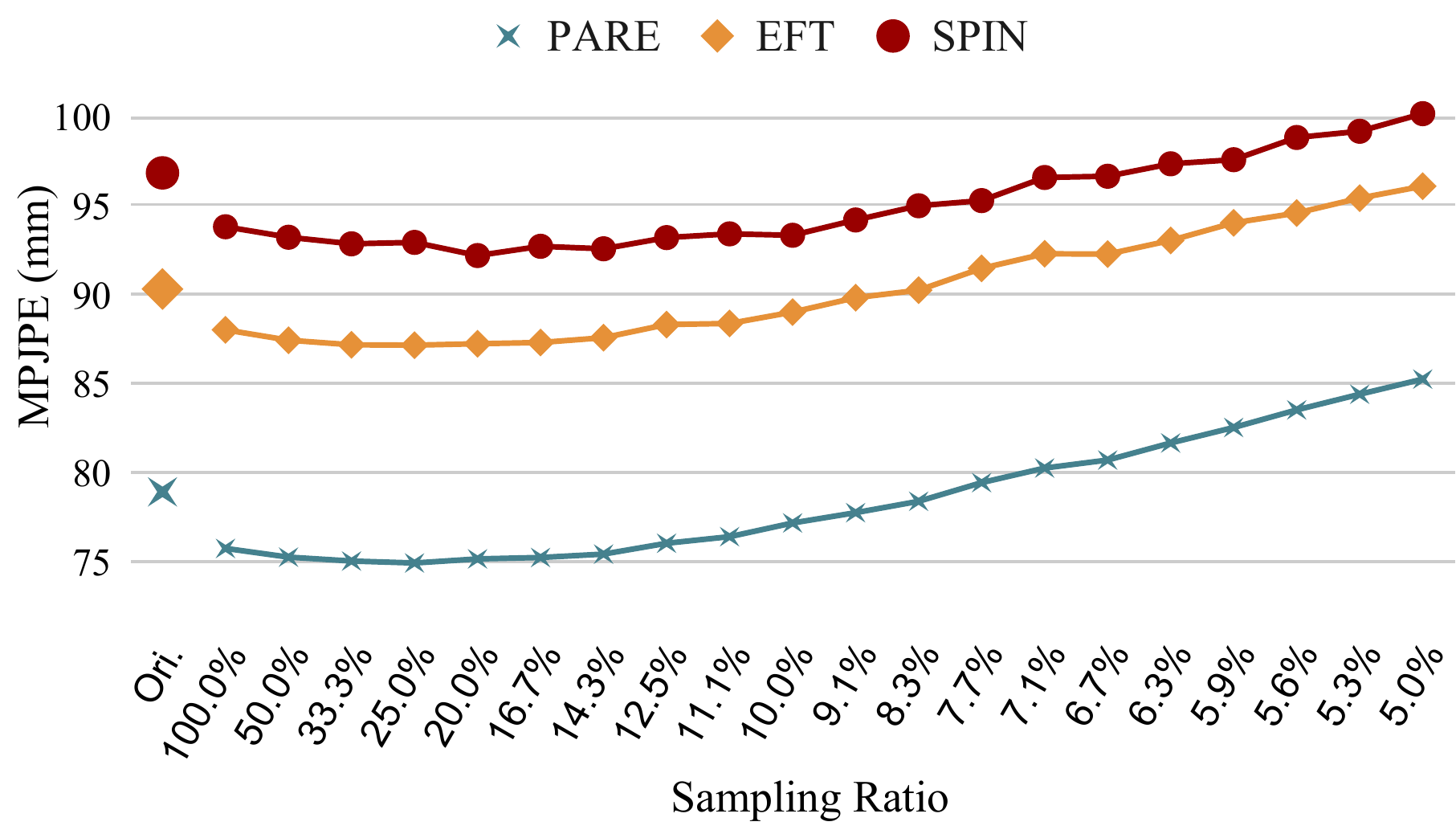}
		\end{minipage}
	}
    	\label{fig:sample_ratio}  
    % 	\hspace{1cm}
	\subfigure[Effect of the window size] 
	{
		\begin{minipage}[t]{0.46\linewidth}
			\centering         
			\includegraphics[width=2.2in]{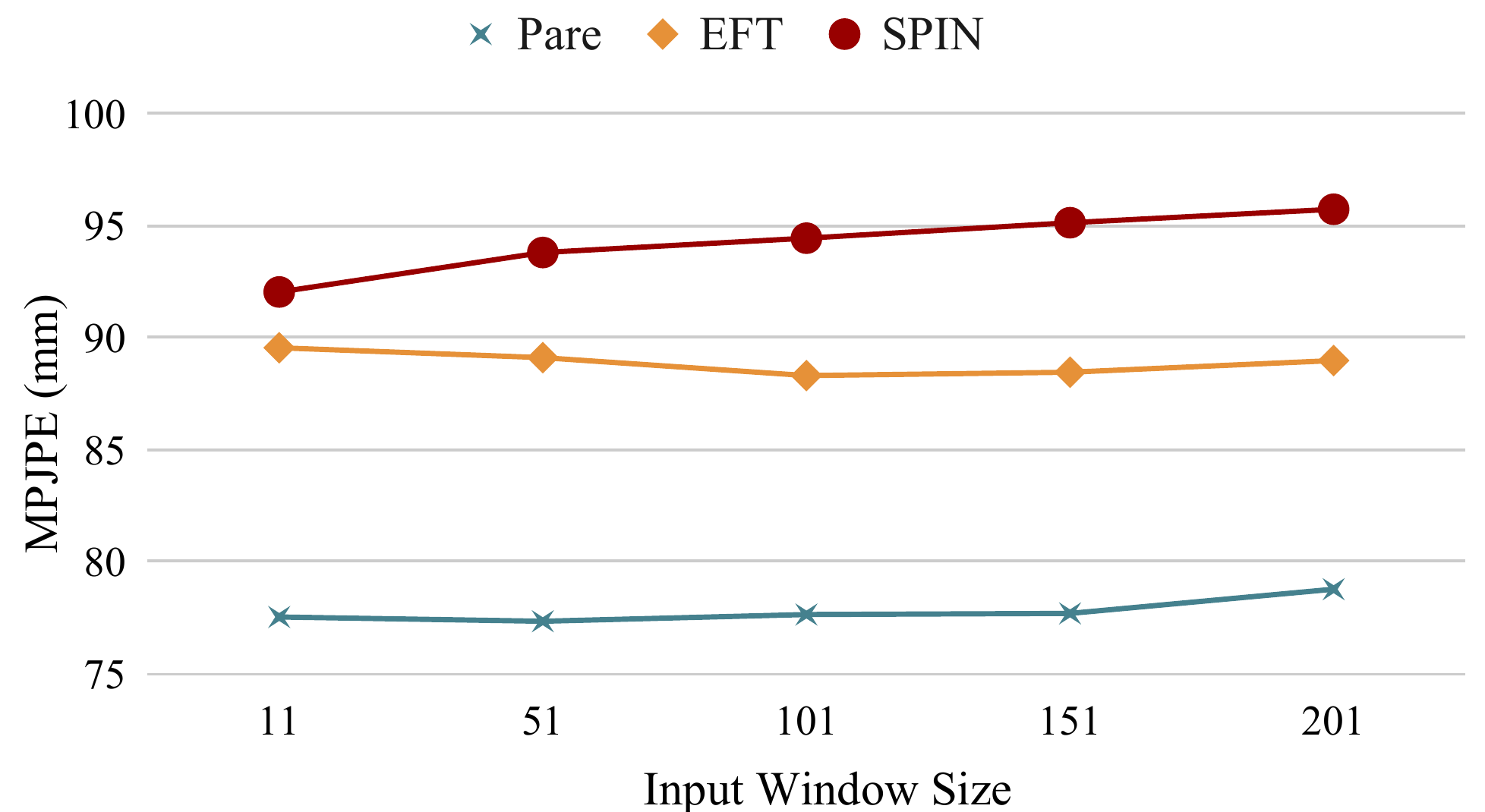}   
		\end{minipage}
	} 
    	\label{fig:window_size} 
    % 	\hspace{1cm}
\caption{\textbf{Comparing effects of different (a) sampling ratios and (b) window sizes.} Sampling interval $N$ is from $1$ to $20$. We compare \emph{MPJPEs} of the three original (Ori.) pose estimators~\cite{kocabas2021pare,joo2020eft,kolotouros2019spin} to our framework on the 3DPW dataset.}
\label{fig:supp_ablation_size} 
%\vspace{-0.5cm}
\end{figure}

\noindent \textbf{Impact of sampling ratio and input window size.}
%W=t*n+1, long coverage
Both input window size and sampling ratio will affect inference efficiency and performance of \name. With the same input window size, the lower the sampling ratio is, the more efficient the inference process will be. We present the comparison of original pose estimator (Ori.)and \name with sampling ratio from $100\%$ to $5\%$ (sampling interval $N$ changes from $1$ to $20$) in Fig.~\ref{fig:supp_ablation_size}(a). When the sampling ratio is $100\%$, \name can be regarded as a denoise model. As shown in Fig.~\ref{fig:supp_ablation_size}(a), the changing trends of \emph{MPJPE} are similar for all three estimation methods (PARE~\cite{kocabas2021pare}, EFT~\cite{joo2020eft}, SPIN~\cite{kolotouros2019spin}). Surprisingly, we find that \emph{MPJPEs} first drop before rising, and they are smallest when the sampling ratio is about $20\%$, with improvements of $4.9\%$, $3.4\%$, and $4.9\%$ for PARE, EFT, and SPIN respectively. This gives us a new perspective that in pose estimation, \emph{not every frame has to be watched to achieve better performance}. The reason behind this is the different degrees of noise in estimated poses. It may be harder to eliminate these diverse degrees of errors in all frames than only denoise some of the frames and recover the rest by temporal continuity.
Besides, the \emph{MPJPEs} of \name is worse than that of the original pose estimator when the sampling ratio is larger than $8\%$ due to too limited input information.

With the sampling ratio fixed at $10\%$, we further explore the influence of window size. 
In Fig.~\ref{fig:supp_ablation_size}(b), we test window sizes from $11$ to $201$. Results indicate that our framework is robust to different window sizes. 

To serve for future research, we report results, including \emph{MPJPEs} and \emph{Accels}, of 3D pose and body estimation on 3DPW, Human3.6M, and AIST++ datasets in Table~\ref{tab:supp_sample_ratio}. All results show similar trends in the change of precision (\emph{MPJPEs}) and smoothness (\emph{Accels}). In addition, \name utilizes the natural smoothness of human motions to recover the detected poses. As a result, the \textit{Accels} decreases steadily when the interval $N$ increases, indicating \name can enhance the smoothness of the existing backbone methods. 

\begin{table}[h]
% \vspace{-0.4cm}
\centering
\caption{\textbf{Results of original (Ori.) estimators~\cite{kocabas2021pare,joo2020eft,kolotouros2019spin,martinez2017simple} and \name under different sampling ratios.} Ori. is the watch-every-frame pose estimator. Sampling interval $N$ is set from $1$ to $20$. The best results are in bold.}
\label{tab:supp_sample_ratio}
\tiny
\setlength{\tabcolsep}{0.7mm}{%
\begin{tabular}{l|ccccccccccccccccccc}
\toprule
Metrics/$N$&  Ori.&1 & 3 & 5&6 &7&8 &9&10&11&12&13&14&15&16&17&18&19&20\\\midrule
\multicolumn{20}{l}{\cellcolor{Gray}\textbf{PARE~\cite{kocabas2021pare} Backbone on 3DPW dataset}}   \\\midrule[0.25pt]                                                              
\emph{MPJPE}&78.9&75.7&\textbf{75.0}&75.1&75.2&75.4&76.0&76.4&77.2&77.7&78.4&79.4&80.3&80.7&81.7&82.5&83.5&84.4& 85.3 \\
\emph{Accel}&25.7&25.2 &9.2&7.7&7.4&7.2&7.1&7.0&6.9&6.9&6.8&6.8&6.7&6.7&6.7&6.7&\textbf{6.6}&\textbf{6.6}&\textbf{6.6}\\\hline
  \multicolumn{20}{l}{\cellcolor{Gray}\textbf{EFT~\cite{joo2020eft} Backbone on 3DPW dataset}} \\\midrule[0.25pt]             
\emph{MPJPE} & 90.3&88.0 &\textbf{87.2} & \textbf{87.2}&87.3 &87.6 & 88.3& 88.4& 89.0& 89.8& 90.3& 91.5&92.3 &92.3 &93.1 &94.0 &94.6 & 95.4&96.1 \\
\emph{Accel} &32.8 &32.7 &10.2 &8.0 &7.5 &7.3 &7.1 &6.9 & 6.8& 6.8& 6.7&6.7 &6.6 &6.6 &6.5 &6.5 &6.5&\textbf{6.4}&\textbf{6.4}
 \\\hline
 \multicolumn{20}{l}{\cellcolor{Gray}\textbf{SPIN~\cite{kolotouros2019spin} Backbone on 3DPW dataset} } \\\midrule[0.25pt]             
\emph{MPJPE} &96.9&93.8&92.9&\textbf{92.2}&92.7&92.6&93.2&93.4&93.3&94.2&95.0&95.3&96.6&96.7&97.4&97.6&98.8&99.2&100.2\\
\emph{Accel}  &34.6 &33.5 &10.5 &8.2 &7.7 & 7.5& 7.3& 7.2& 7.1&7.0 &6.9 &6.9 &6.9 &6.9 &\textbf{6.8} &6.9 &\textbf{6.8}&\textbf{6.8} &\textbf{6.8}
 \\\toprule \hline
 \multicolumn{20}{l}{\cellcolor{Gray}\textbf{FCN~\cite{martinez2017simple} Backbone on Human3.6M dataset} } \\\midrule[0.25pt]         
\emph{MPJPE}  &54.6 &53.3 & 53.0& 52.8& 52.6&\textbf{52.3} &52.5 &52.6 &52.8 &53.0 & 53.2& 53.2& 53.4& 53.5&53.9 &53.8 &54.0 &54.2 &54.4 \\
\emph{Accel}  &19.2 &15.4 &3.1 &2.0 &1.8 &1.6 &1.6 &1.5 &1.5 &\textbf{1.4} &\textbf{1.4} &\textbf{1.4} &\textbf{1.4} &\textbf{1.4} & \textbf{1.4}&\textbf{1.4} &\textbf{1.4} &\textbf{1.4} &\textbf{1.4}
 \\\toprule \hline
  \multicolumn{20}{l}{\cellcolor{Gray}\textbf{SPIN~\cite{kolotouros2019spin} Backbone on AIST++ dataset} }  \\\midrule[0.25pt]
\emph{MPJPE}  & 107.7&67.2 &\textbf{66.6} &67.6 &68.4 &69.7 &71.2 &71.6 &71.3 &76.1 &77.1 &79.0 &80.2 &82.3 &84.3 &85.2 &87.0 &88.9 &90.8 \\
\emph{Accel}  & 33.8& 7.6& 7.6&6.6 &6.3 &6.1 &6.0 &5.9 &5.7 &5.7 &5.6 &5.6 &5.5 &5.5 &5.5 &5.5 &5.4 &\textbf{5.3} &\textbf{5.3} 
 \\\bottomrule
\end{tabular}%
}
%\vspace{-0.5cm}
\end{table}

\noindent \textbf{Analyses on the phenomenon: fewer samples with better performance.}
When the inputs of \name are \underline{ground-truth poses}, the performance deteriorates with decreased sampling ratio (see Table~\ref{tab:label} above). However, in practice, the inputs of \name are \underline{noisy detected poses}, and some of them have high errors (e.g., due to occlusion). 
Consequently, considering the detected poses' errors, two factors affect the recovered poses.

1) On the one hand, not considering/aggregating the highly noisy poses can improve performance by reducing the impact of noisy poses on both \emph{DenoiseNet} and \emph{RecoverNet}. 2) On the other hand, dropping too many frames would lead to performance degradation due to information insufficiency.

Generally speaking, when the sampling ratio is high (e.g., \textgreater 20\%), we could easily recover intermediate poses thanks to the continuity of motions.
And for those dropped intermediate poses,
the denoised and recovered poses via \name obtain lower error compared to their original noisy estimation.
Consequently, the overall MPJPEs would drop with the increase of intervals (i.e., the decrease of sampling ratio) in the beginning. However, when the sampling ratio becomes too low (e.g., \textless 5\%), the highly sparse poses do not provide sufficient information for motion recovery, and the MPJPEs would go up under such circumstances. In other words, there would be a ``sweet spot'' for the sampling ratio with the minimum MPJPEs.  
This phenomenon is also present in the traditional interpolation method, where \emph{MPJPE} first drops (from 107.7mm to 105.8mm) before rising.

\begin{table}[H]
\centering
\tabcolsep=8pt
 
    \caption{Comparison results with different denoise network designs on 3DPW dataset with the state-of-the-art pose estimator Pare~\cite{kocabas2021pare} (\emph{MPJPE} is $78.9$mm).}
	{%
		\begin{tabular}{l|c|c|c|c}
			\specialrule{.1em}{.05em}{.05em}
			
			Metrics& No \emph{DenoiseNet} & TCNs~\cite{pavllo20193d} & MLPs~\cite{zeng2021smoothnet} & Ours\\
			\midrule
		    \emph{MPJPE}&79.8&80.5&79.5&\textbf{77.2}\\
        \midrule
        \end{tabular}%
	}
	\label{tab:denoise}
\end{table}

\noindent \textbf{Study on different denoise networks.}
%spatial-temporal
As a baseline framework, we try to explore the performance of different network designs of the two subnets, \textit{DenoiseNet} and \textit{RecoverNet}. In the second step, we use \emph{DenoiseNet} with Transformer architecture to relieve noises from single-frame estimators. We first remove this network to validate its effectiveness. Table~\ref{tab:denoise} shows a $2.1$mm reduction of \emph{MPJPE} without \emph{DenoiseNet}, indicating this step is essential to the recovery of more precise poses. Then, we try to simply replace the Transformer with TCNs~\cite{pavllo20193d}, with zero paddings to make the input and output length the same, and MLPs~\cite{zeng2021smoothnet} along temporal axes. Results show these models are incapable of handling the discrete diverse noises, making the final recovery results worse than the original result.

%linear 
\begin{table}[H]
\tabcolsep=8pt
	\centering
    \caption{\textbf{Comparison results with different recovery network designs} on 3DPW dataset with the state-of-the-art pose estimator PARE~\cite{kocabas2021pare}(\emph{MPJPE} is $78.9$mm).}
	{%
		\begin{tabular}{l|c|c|c|c|c}

			\specialrule{.1em}{.05em}{.05em}
			
			Metrics&Linear&TCNs~\cite{pavllo20193d} &TCNs w/MLPs& MLPs~\cite{zeng2021smoothnet} &Ours\\
			\midrule
		    \emph{MPJPE}&79.8& 172.3&99.5&78.0&\textbf{77.2}\\
        \midrule
        \end{tabular}%
	}
	\label{tab:recovery}
\end{table}

\noindent \textbf{Study on different recovery methods.}
Lastly, we analyze possible designs of the recovery process in the third step. First, we try the simple Linear interpolation, which shows more significant errors compared with the original PARE since it loses the non-linear motion dynamics. Then, we adopt TCNs~\cite{pavllo20193d}, which have local temporal receptive fields (e.g., $3$) in each layer to recover the missing values with the interval $N$ as $10$, and it leads to the worst results. After adding MLPs~\cite{zeng2021smoothnet} at the last layer to enhance long-term temporal coherence, the error reduces from $172.3$mm to $99.5$mm (by $42.3$\% improvement), but the error is still far from satisfactory. MLPs~\cite{zeng2021smoothnet} can utilize the continuity of temporal dimension to learn non-linear fitting curves from sampled points. Still, they do not aggregate spatial information, which makes them get a slightly worse result. 
%We also attempt to use the 1D masked autoencoder (MAE~\cite{he2021masked}) and Interpolation network~\cite{ho2021render}, while obtaining slightly unsatisfactory performance compared with \emph{RecoverNet}. 
% Note that the key differences from existing pose transformers is the temporal semantic tokens to bring atomic motion knowledge into spatio-temporal correlation learning. 

\begin{table}[H]
\tabcolsep=8pt
\centering
    \caption{\textbf{Comparison results of different loss weight $\lambda$} on 3DPW dataset with the state-of-the-art pose estimator Pare~\cite{kocabas2021pare}(MPJPE is $78.9$mm).}
	{%
		\begin{tabular}{l|c|c|c|c}

			\specialrule{.1em}{.05em}{.05em}
			
			$\lambda$& 1 & 2 & 5 & 10\\
			\midrule
		    \emph{MPJPE}&78.0&77.6&\textbf{77.2}&77.5\\
        \midrule
        \end{tabular}%
	}
	\label{tab:supp_loss}
\end{table}

\noindent \textbf{Study on different hyper-parameters.}
We also show the effects of hyper-parameters in \name. 
$\lambda$ is used in the loss function to balance the losses between \emph{RecoverNet} and \emph{DenoiseNet}. Results in Table~\ref{tab:supp_loss} show that \emph{MPJPEs} are robust to diverse loss values. Therefore, we set it to $5$ by default.
Moreover, we use the same embedding dimension $C$ and block number $M$ in transformer blocks. We show the results of different $C$ in Table~\ref{tab:supp_c} and $M$ in Table~\ref{tab:supp_m}. Fewer parameters, such as $C=12$ and $M=1$, will lead to performance degradation. As the model becomes deeper (larger $M$) and wider (larger $C$), the performance will meet saturation. By default, we set $C=64$ and $M=5$ for all experiments.

\begin{table}[H]
\tabcolsep=8pt
    \centering
    \caption{\textbf{Comparison results of different embedding dimension $C$} on 3DPW dataset with the state-of-the-art  pose estimator Pare~\cite{kocabas2021pare}.}
	{%
		\begin{tabular}{l|c|c|c|c|c}

			\specialrule{.1em}{.05em}{.05em}
			
			$C$& 12 & 32 & 64& 128&256\\
			\midrule
		    \emph{MPJPE}&78.0&\textbf{77.2}&77.4&77.6&77.4\\
        \midrule
        \end{tabular}%
	}
	\label{tab:supp_c}
	\centering
\end{table}

\begin{table}[H]
\centering
\tabcolsep=8pt
    \caption{\textbf{Comparison results of different block number $M$} on 3DPW dataset with the state-of-the-art pose estimator Pare~\cite{kocabas2021pare}.}
	{%
		\begin{tabular}{l|c|c|c|c}

			\specialrule{.1em}{.05em}{.05em}
			
			$\lambda$& 1 & 3 & 5 & 10\\
			\midrule
		    \emph{MPJPE}&79.3&77.5&\textbf{77.2}&77.6\\
        \midrule
        \end{tabular}%
	}
	\label{tab:supp_m}
\end{table}

\section{Qualitative Results}
\label{sec:supp_viz}
We demonstrate three typical successful cases of \name to understand why \name uses fewer frames with higher efficiency but gets better performance than existing single-frame methods.

\begin{figure}[h]	
\centering
 	
 		\begin{minipage}[t]{0.98\linewidth}
 			\centering         
 			\includegraphics[width=4.5in]{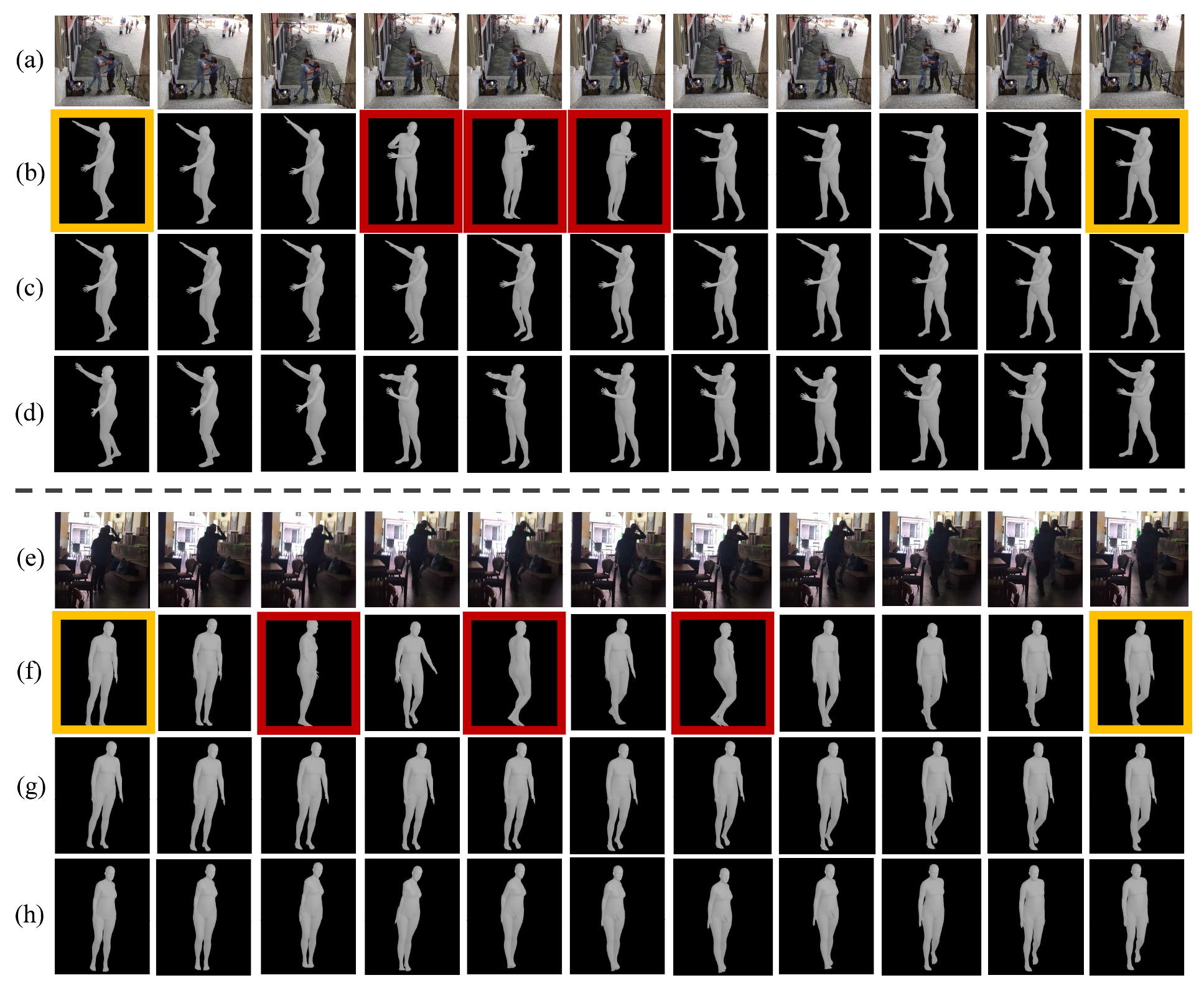}   
 		\end{minipage}

% \vspace{-0.1cm}
\caption{Visualization results of estimated body recovery from two video sequences with eleven frames in (a) and (e) rows. (b) and (f) are estimated poses from the existing SOTA model PARE~\cite{kocabas2021pare}. We highlight the input poses of \name in the yellow boxes and the high-error poses in the red boxes. (c) and (g) are output poses of our proposed \name, the sampling ratio is 10\% in this framework. (d) and (h) show the ground truth of the corresponding poses. }
\label{fig:supp_viz_1} 
%\vspace{-0.4cm}
\end{figure}

% \noindent \textbf{Visualization on how precision is improved.}
First, cases in Fig.~\ref{fig:supp_viz_1} show that \name can improve not only efficiency but also effectiveness on the 3D body recovery task. The estimated body in the yellow boxes are inputs of \name, where the interval $N$ is set to $10$. Existing SOTA models, like PARE~\cite{kocabas2021pare}, will fail (illustrated in red boxes) when the frames have heavy body occlusions, human interactions, or poor image quality. Interestingly, \name skips some frames (inputs are in yellow boxes) to avoid the negative effect. Therefore, compared with the watch-every-frame model~\cite{kocabas2021pare}, \name may reduce the effects of unreliable and noisy estimated poses by a temporal recovery scheme to obtain the rest of the results. 

\begin{figure}[h]	
\centering
 	
 		\begin{minipage}[t]{0.98\linewidth}
 			\centering         
 			\includegraphics[width=4.5in]{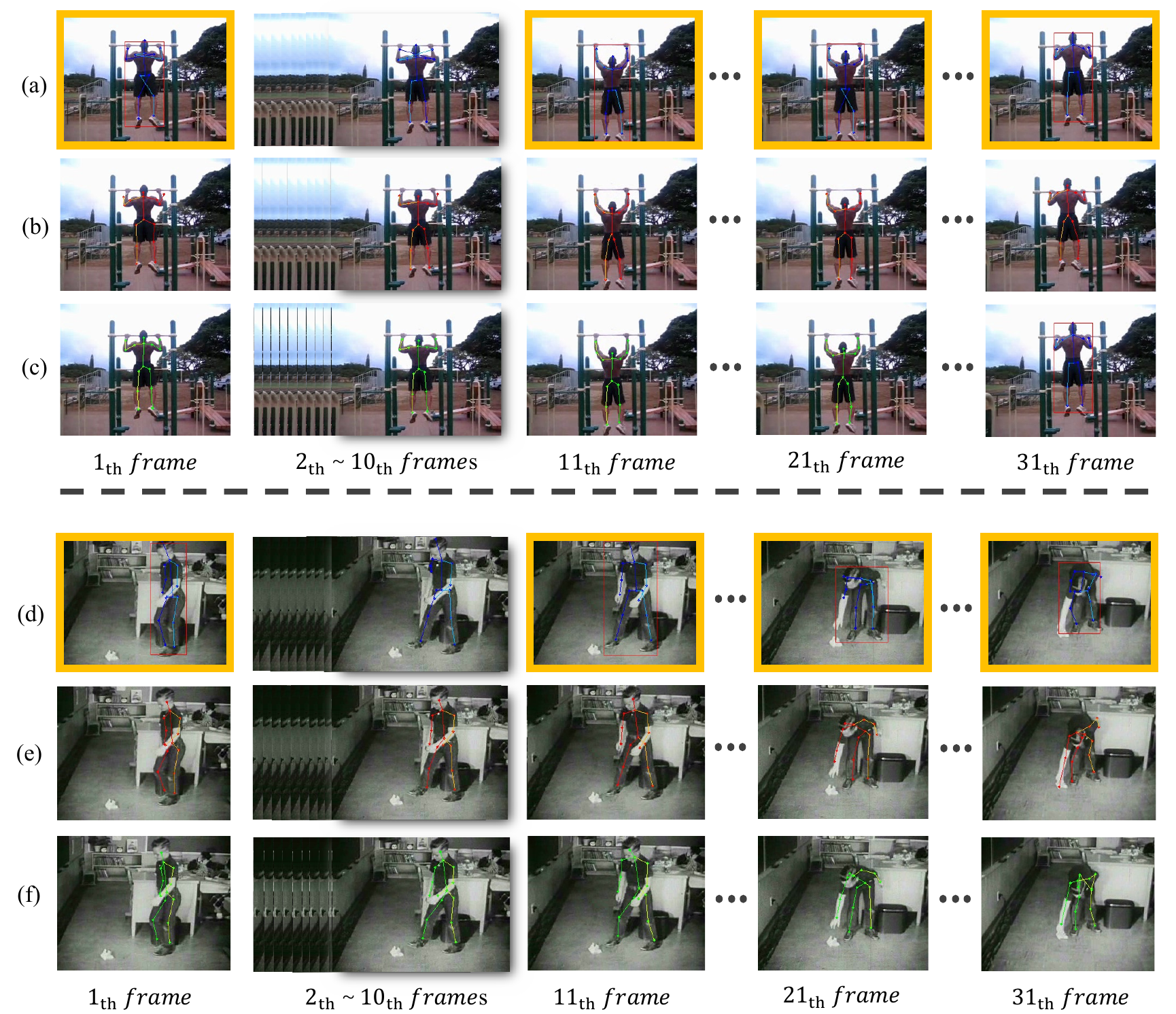}   
 		\end{minipage}     	
    	
% \vspace{-0.1cm}
\caption{Visualization the impact of denoising scheme in \name on calibrating the wrong detected poses on four visible frames from the single-frame backbone. We demonstrate the cases via two video sequences and simply ignore the invisible frames. (a) and (d) rows show estimated poses from the popular model SimplePose~\cite{xiao2018simple}, where the sampling interval is 10. Inputs of \name are highlighted in the yellow boxes. (b) and (e) are  output poses of our proposed \name, which can denoise and smooth the input poses by the proposed \emph{DenoiseNet} and \emph{RecoverNet}. (c) and (f) show the ground truth of the corresponding poses. }
\label{fig:supp_viz2} 
%\vspace{-0.4cm}
\end{figure}

\emph{What happens if there are mistakes in the visible frames?} In Fig.~\ref{fig:supp_viz2}, we show the impact of denoising scheme in \name on 2D pose estimation. Given a sliding window of $31$ frames, we mainly demonstrate the visible four frames (highlighted in yellow boxes) with their detected 2D poses by SimplePose~\cite{xiao2018simple}. We observe that there are left-right flipped keypoint detection in the $1_{th}$ and $21_{th}$ frames of Fig.~\ref{fig:supp_viz2}(a), which sometimes happens when the input image is the back of the person. In the $31_{th}$ frame of Fig.~\ref{fig:supp_viz2}(d), high errors occur due to heavy self-occlusion. Our method utilizes long temporal effective receptive fields to denoise the noisy input poses and then recover the clean sparse poses to get the final sequence poses, making the output poses smooth and precise in an efficient way.

\begin{figure}[h]	
\centering
 	
 		\begin{minipage}[t]{0.98\linewidth}
 			\centering         
 			\includegraphics[width=4.5in]{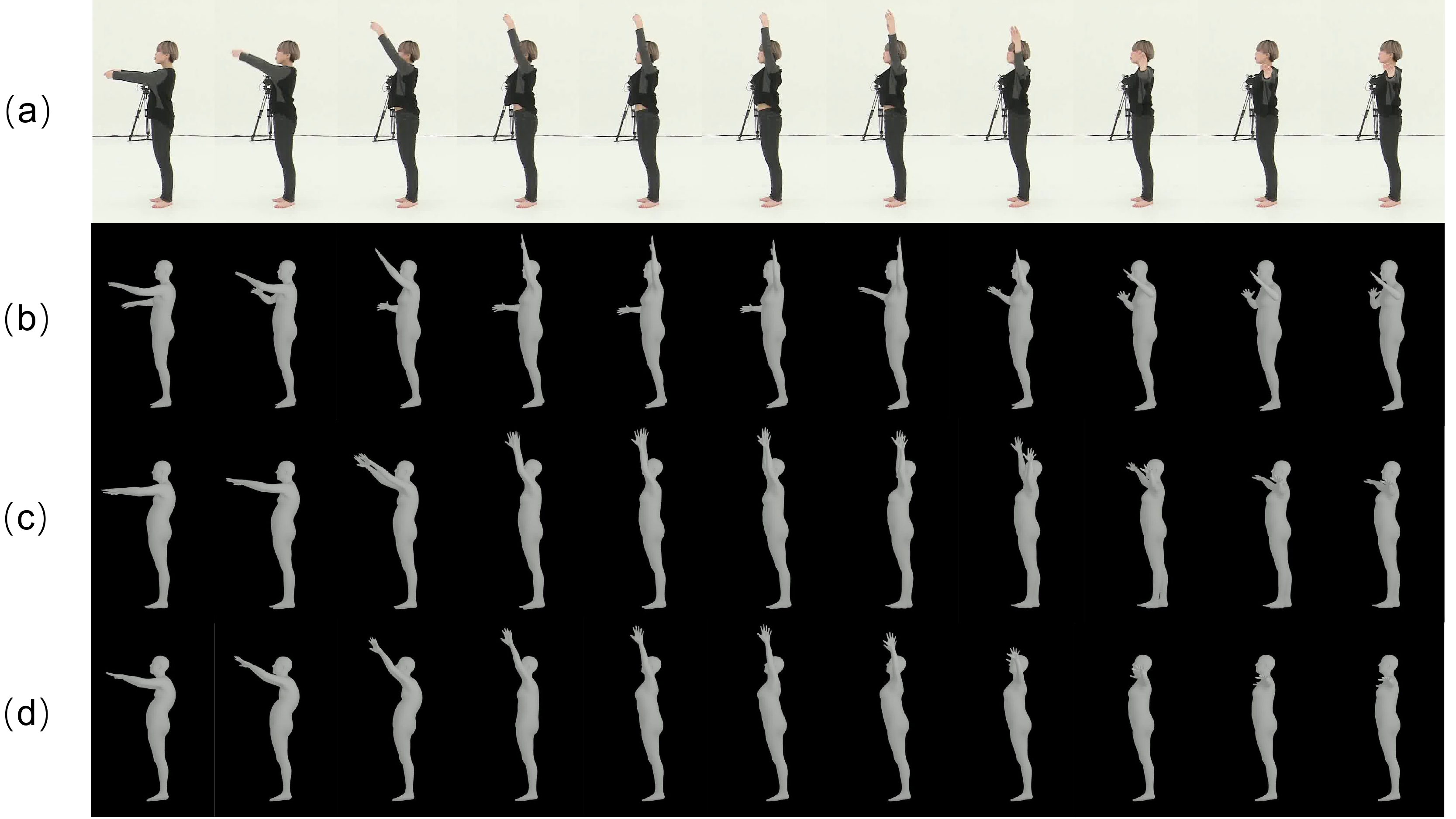}   
 		\end{minipage}
    	
% \vspace{-0.1cm}
\caption{Visualization of the recovery results on high-error estimated poses from AIST++ dataset. Only visible frames are shown, which are sampled with an interval of $10$. Images in the row (a) are the original input frames at the $1_{th}$, $11_{th}$, $21_{th}$,...,$101_{th}$ frame. (b), (c), (d) show the poses detected by SPIN~\cite{kolotouros2019spin}, poses recovered by \name, and the corresponding ground truth. }
\label{fig:all_wrong_visible} 
%\vspace{-0.4cm}
\end{figure}

\emph{In addition to being able to do better motion sequence recovery, can \name still learn motion prior?}
In some cases, even if all visible frames are inaccurate, \name can still recover accurate poses by learning motion prior. As shown in Fig.~\ref{fig:all_wrong_visible}, (a) shows the original video frames of AIST++ with an interval of $10$, which is all the visible frames in one slide window (a sliding window with the length of $101$ has $11$ visible frames). (b) is the corresponding SMPL pose detected by SPIN~\cite{kolotouros2019spin}. Large errors occur in the actor's occluded right arm and hand. In Fig.~\ref{fig:all_wrong_visible}(c), \name can successfully correct the errors and outputs smooth poses leveraging dancing action prior and human motion continuity, which are hard for existing single-frame estimators to estimate occluded body parts. (d) shows the ground truth poses of the video frames.

For more visualization of 2D pose estimation, 3D pose estimation as well as body recovery, please refer to our website\footnote{Website: \url{https://ailingzeng.site/deciwatch}}.

\section{Failure Case Analyses}
\label{sec:supp_fail}

There are two types of failure cases in \name, which motivates the two corresponding future directions. 
\begin{itemize}

\item \emph{When the sampling rate is lower than the motion frequency of some body parts, it will be difficult to supplement the actual motion.} Human body is articulated. Thus different body parts have different movement frequencies and distribution. For example, the moving frequency and amplitude of hands and feet will be greater than that of the trunk. Our method adopts the same sampling rate for the whole body without considering that the motion distribution of different keypoints is different. In some actions, such as playing the guitar, only the hand will move at high frequency, but most other joints will not move, so the detailed information recovery of hand movement will be lost. Therefore, adaptive sampling strategies, especially on different body parts or keypoints, will be beneficial.

\item \emph{If the estimated poses of most visible frames in the sliding window are in large errors, it is hard for \name to recover the correct poses.} As shown in Fig.~\ref{fig:supp_viz2}, although our method can correct the noisy poses to some extent, this is the advantage of learnable methods. That is, the traditional interpolation method can not fix them. However, if most of the visible poses are noisy, our output may also tend to have similar (but smooth) errors. Thus, it is still essential to continuously improve the performance and robustness of pose estimation methods, especially in extreme scenes. At the same time, we can also consider using additional lightweight information, such as IMUs, to help improve performance.

\end{itemize}

% \noindent \textbf{Comparison of motion completion methods.}

% \begin{figure}[h]	
% \centering
 	
%  		\begin{minipage}[t]{0.98\linewidth}
%  			\centering         
%  			\includegraphics[width=4.5in]{figures/aist_spin_deci_0.1_all_seq1174.pdf}   
%  		\end{minipage}
%  		\label{fig:supp_viz1}

% % \vspace{-0.1cm}
% \caption{Comparison with the most used motion completion methods, linear interpolation and quadratic interpolation for both position errors and acceleration errors.}
% \label{fig:supp_viz2} 
% %\vspace{-0.4cm}
% \end{figure}

% \noindent \textbf{Visualization on failure cases.}

\clearpage

\bibliographystyle{splncs04}